# From pixels to planning: scale-free active inference

Karl Friston[1,2], Conor Heins[2], Tim Verbelen[2], Lancelot Da Costa[1,2], Tommaso Salvatori[2], Dimitrije Markovic[2,3], Alexander Tschantz[2], Magnus Koudahl[2], Christopher Buckley[2,4], Thomas Parr[5].

[1] Queen Square Institute of Neurology, University College London, UK
[2] VERSES Research Lab, Los Angeles, California, 90016, USA
[3] Chair of Cognitive Computational Neuroscience, Technische Universität, Dresden, Germany
[4] Department of Informatics, University of Sussex, Brighton, UK
[5] Nuffield Department of Clinical Neurosciences, University of Oxford, UK

Emails: k.friston@ucl.ac.uk; conor.heins@verses.ai; tim.verbelen@verses.ai; lance.dacosta@verses.ai; tommaso.salvatori@verses.ai; dimitrije.markovic@tu-dresden.de; alec.tschantz@verses.ai; c-magnus.koudahl@verses.ai; christopher.buckley@verses.ai; thomas.parr@ndcn.ox.ac.uk;

## ABSTRACT

This paper describes a discrete state-space model—and accompanying methods—for generative modelling. This model generalises partially observed Markov decision processes to include paths as latent variables, rendering it suitable for active inference and learning in a dynamic setting. Specifically, we consider deep or hierarchical forms using the renormalisation group. The ensuing *renormalising generative models* (RGM) can be regarded as discrete homologues of deep convolutional neural networks or continuous state-space models in generalised coordinates of motion. By construction, these scale-invariant models can be used to learn compositionality over space and time, furnishing models of paths or orbits; i.e., events of increasing temporal depth and itinerancy. This technical note illustrates the automatic discovery, learning and deployment of RGMs using a series of applications. We start with image classification and then consider the compression and generation of movies and music. Finally, we apply the same variational principles to the learning of Atari-like games.

**Keywords**: *active inference; active learning; Bayesian model selection; renormalisation group; compression; structure learning; RG-flow*

## INTRODUCTION

This paper considers the use of discrete state-space models as generative models for classification, compression, generation, prediction and planning. The inversion of such models can be read as inferring the latent causes of observable outcomes or content. When endowed with the consequences of action, they can be used for planning as inference (Attias, 2003; Botvinick and Toussaint, 2012; Da Costa et al., 2020). This allows one to cast any classification, prediction, or planning problem as an inference problem that—under active inference—reduces to maximising model evidence. However, applications of active inference have been largely limited to small-scale problems. In this paper, we consider one solution to the implicit scaling problem; namely, the use of scale-free generative models and the renormalisation group (Cardy, 2015; Cugliandolo and Lecomte, 2017; Hu et al., 2020; Lin et al., 2017; Schwabl, 2002; Vidal, 2007; Watson et al., 2022). The contribution of this paper is to consider *discrete* models that are renormalisable in state-space *and time*.

Specifically, this paper explores the use of generalised Markov decision processes as discrete models apt for compressing data, generating content or planning. The generalisation in question rests on equipping a standard (partially observed) Markov decision process with random variables called *paths*. This affords an expressive model of dynamics, in which transitions among states are conditioned upon paths, which themselves can have lawful transitions. This generalisation becomes particularly important when composing Markovian processes in a deep or hierarchical architecture; e.g., (Friston et al., 2017b). This follows from the use of states at one level to generate the initial conditions and paths at a lower level. In effect, this means that states generate paths, which generate states, which generate paths, and so on; furnishing trajectories with deep, semi-Markovian structure: e.g., (Marković et al., 2022). Their recursive aspect speaks to a definitive feature of the generative models considered in this paper—renormalisability.

Intuitively, renormalisability rests on a renormalisation group (RG) operator, which takes a description of the system at hand (e.g., an action, partition function, *etc.*), and returns a coarse-grained version that retains the properties of interest, while discarding irrelevant details (Cardy, 2015; Schwabl, 2002; Watson et al., 2022). For excellent overviews of the renormalisation group in machine learning, please see (Hu et al., 2020) and (Mehta and Schwab, 2014). See also (Lin et al., 2017) who appeal to the renormalisation group to formalise the claim that: "when the statistical process generating the data is of a certain hierarchical form prevalent in physics and machine learning, a deep neural network can be more efficient than a shallow one." Crucially, any random dynamical system with sparse coupling—and an implicit Markov blanket partition (Sakthivadivel, 2022)—is renormalisable (Friston, 2019). Therefore, any generative model that recapitulates 'the statistical process generating the data' must be renormalisable.

In what follows, we illustrate applications of the same *renormalising generative model* (RGM) in several settings. The universality of this model calls upon the apparatus of the renormalisation group. In brief, its deep structure ensures that each level can be renormalised to furnish the level above. The renormalisation group requires that the functional form of the dynamics (e.g., belief updating) is conserved over levels or scales. This is assured in variational inference; in the sense that the inference process itself can be cast as pursuing a path of least action (Friston et al., 2023a), where action is the path integral of variational free energy: c.f., (Mehta and Schwab, 2014). The only thing that changes between levels are the parameters of the requisite action (e.g., sufficient statistics of various probability distributions). The relationship between the parameters at one level and the next rests on an RG operator that entails a grouping and dimension reduction, i.e., coarse graining or scaling transformation. By choosing the right kind of RG operator, one can effectively dissolve the scaling problem. In short, by ensuring each successive level of a deep generative model is renormalisable, one can—in principle—generate data at any scale. And, implicitly, infer or learn the causes of those data. The notion of scale invariance is closely related to universality, licensing the notion of a universal generative model.

Instances of the renormalisation group abound in natural and machine learning: for example, the cortical visual hierarchy in the brain, with progressive enlargement of spatiotemporal receptive fields as one ascends hierarchical levels or scales (Angelucci and Bullier, 2003; Hasson et al., 2008; Zeki and Shipp, 1988). The same kind of architectures—associated with deep convolutional neural networks—could almost be definitive of deep models and learning (Hu et al., 2020; Lin et al., 2017). Here, we pay special attention to the implications of universality and scale-invariance for specifying the structural form of generative models and illustrate the ensuing efficiency when deployed in some typical use cases.

This paper comprises four sections. The first rehearses the variational procedures or methods used in active inference, learning and selection, with a special focus on the selection of hierarchical model structures that can be renormalised. In brief, active *inference*, *learning* and *selection* speaks to the distinct sets of unknown variables that constitute a generative model; namely, latent *states*, *parameters* and *structure*. On this view, model inversion corresponds to (Bayesian) belief updating at each of these levels, by minimising *variational free energy*; i.e., maximising an evidence lower bound (Winn and Bishop, 2005). The active part of inference, learning and selection arises operationally through selecting or choosing those actions that minimise *expected free energy*, which can be decomposed in number of ways that subsume commonly used objective functions in statistics and machine learning (Da Costa et al., 2020). The dénouement of this section considers the structural form of renormalising architectures, illustrated by successive elaborations in the remaining sections.

The second section starts with a simple application to models of static images that can be read as a form of image compression; i.e., maximising model evidence via minimising model complexity through the compression afforded by successive block-spin transformations (Vidal, 2007). The implicit sample efficiency is showcased by application to the MNIST digit classification problem. This application foregrounds the representational nature of the generative model, moving from a *place-coded* representation at lower levels to an *object-centred* representation (i.e., digit classes) at the highest level. The next section uses the same methods to illustrate renormalisation over time, i.e., modelling paths or sequences of increasing temporal depth at successively higher levels. This application can be regarded as a form of video compression—illustrated using short movies files—that can be used to recognise sequences of events or generate sequences in response to a prompt. In contrast to the object-centred representations—afforded by application to static images—this section speaks to *event-based* compression; suitable for classifying or generating visual or auditory scenes. The next section leverages the ability to classify or generate sequences by applying the same methods to sound files, illustrated using birdsong and music. The final section turns to planning and agency by using an RGM to learn and play (Atari-like) games. This application involves equipping the generative model with the capacity to act; namely, to realise the predicted consequences of action, where these predictions are based upon a fast form of structure learning, effectively evincing a one-shot learning of expert play.

Although the focus on renormalisation inherits from the physics of universal phenomena (Schwabl, 2002; Vidal, 2007), we highlight the biomimetic aspects of inference and learning that emerge under these models. The implication here is that natural intelligence may have evolved renormalising structures just because the world features universal phenomena, such as scale-invariance. This is not a machine learning paper because the objective in active inference is to maximise model evidence. We therefore refrain from benchmarking any of the examples in terms of performance or accuracy. However, it should be self-evident that the methods on offer are generally more sample efficient than extant (machine) learning schemes[1].

# ACTIVE INFERENCE, LEARNING AND SELECTION

This section overviews the model used in the numerical studies of subsequent sections. This model generalises a partially observed Markov decision process (POMDP) by equipping it with random variables called *paths* that 'pick out' dynamics or transitions among latent states. These models are designed to be composed hierarchically, in a way that introduces a separation of temporal scales.

## Generative models

Active inference rests upon a *generative model* of observable outcomes. This model is used to infer the most likely causes of outcomes in terms of expected states of the world. These states (and paths) are latent or *hidden* because they can only be inferred through observations. Some paths are controllable, in that they can be realised through action. Therefore, certain observations depend upon action (e.g., where one is looking), which requires the generative model to entertain expectations about outcomes under different combinations of actions (i.e., policies)[2]. These expectations are optimised by minimising *variational free energy*. Crucially, the prior probability of a policy depends upon its *expected free energy*. Having evaluated the expected free energy of each policy, the most likely action is selected and the implicit perception-action cycle continues (Parr et al., 2022).

---

[1] To avoid overburdening the main text, some details of the numerical studies have been placed in figure legends.

Figure 1 provides an overview of the generative model considered in this paper. Outcomes at any time depend upon hidden *states*, while transitions among hidden states depend upon *paths*. Note that paths are random variables, that may or may not depend upon action. The resulting POMDP is specified by a set of tensors. The first set **A**, maps from hidden states to outcome modalities; for example, exteroceptive (e.g., visual) or proprioceptive (e.g., eye position) *modalities*. These parameters encode the likelihood of an outcome given their hidden causes. The second set **B** encodes transitions among the hidden states of a *factor*, under a particular path. Factors correspond to different kinds of causes; e.g., the location versus the class of an object. The remaining tensors encode prior beliefs about paths **C**, and initial conditions **D** and **E**. The tensors are generally parameterised as Dirichlet distributions, whose sufficient statistics are concentration parameters or *Dirichlet counts*, which count the number of times a particular combination of states and outcomes has been inferred. We will focus on learning the likelihood model, encoded by Dirichlet counts, *a*.

---

[2] In this setting a policy is not a sequence of actions but a combination of paths, where each hidden factor has an associated state and path. This means there are, potentially, as many policies as there are combinations of paths.

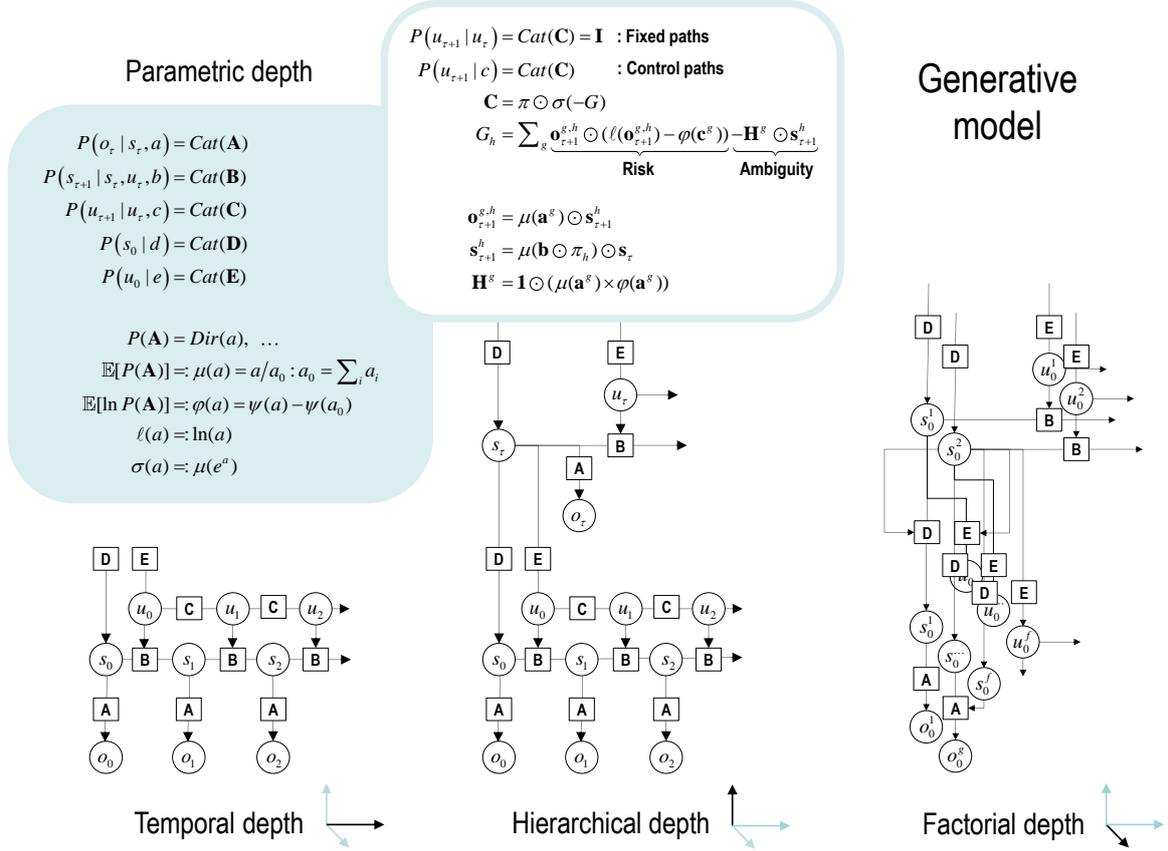

**Figure 1: Generative models.** A generative model specifies the joint probability of observable consequences and their hidden causes. Usually, the model is expressed in terms of a *likelihood* (the probability of consequences given their causes) and *priors* (over causes). When a prior depends upon a random variable it is called an *empirical prior*. Here, the likelihood is specified by a tensor **A**, encoding the probability of an outcome under every combination of *states* (*s*). Priors over transitions among hidden states, **B** depend upon *paths* (*u*), whose transition probabilities are encoded in **C**. Certain (*control*) paths are more probable *a priori* if they minimise their expected free energy (**G**), expressed in terms of *risk* and *ambiguity* (white panel). If the path is not controllable, it remains *fixed* over the epoch in question, where **E** specifies the prior over paths. The left panel provides the functional form of the generative model in terms of categorical (*Cat*) distributions that are themselves parameterised as Dirichlet (*Dir*) distributions, equipping the model with the *parametric depth*. The lower equalities list the various operators required for variational message-passing in Figure 2. These functions are taken to operate on each column of their tensor arguments. The graph on the lower left depicts the generative model as a probabilistic graphical model that foregrounds the implicit *temporal depth* implied by priors over state transitions and paths. This example shows dependencies for fixed paths. When equipped with *hierarchical depth* the POMDP acquires a separation of temporal scales. This follows because higher states generate a sequence of lower states, i.e., the initial state (via the **D** tensor) and subsequent path (via the **E** tensor). This means higher levels unfold more slowly than lower levels, furnishing empirical priors that contextualise the dynamics of their children. At each hierarchical level, hidden states and accompanying paths are factored to endow the model with *factorial depth*. In other words, the model 'carves nature at its joints' into factors that interact to generate outcomes (or initial states and paths at lower levels). The implicit context-sensitive contingencies are parameterised by tensors mapping from one level to the next (**D** and **E**). Subscripts pertain to time, while superscripts denote distinct factors (*f*), outcome modalities (*g*) and combinations of paths over factors (*h*). Tensors and matrices are denoted by uppercase bold, while posterior expectations are in lowercase bold. The matrix $\pi$ encodes the probability over paths, under each *policy* (for notational simplicity, we have assumed a single control path). The $\odot$ notation implies a generalised inner (i.e., dot) product or tensor contraction, while × denotes the Hadamard (element by element) product. $\psi(\cdot)$ is the digamma function, applied to the columns of a tensor.

The generative model in Figure 1 means that outcomes are generated as follows: first, a policy is selected using a softmax function of expected free energy. Sequences of hidden states are generated using the probability transitions specified by the selected combination of paths (i.e., policy). Finally, these hidden states generate outcomes in one or more modalities. *Inference* about hidden states (i.e., state estimation) corresponds to inverting a generative model, given a sequence of outcomes, while *learning* corresponds to updating model parameters. The requisite expectations constitute the sufficient statistics $(\mathbf{s}, \mathbf{u}, \mathbf{a})$ of posterior beliefs $Q(s, u, a) = Q_\mathbf{s}(s) Q_\mathbf{u}(u) Q_\mathbf{a}(a)$. The implicit factorisation of this approximate posterior effectively partitions model inversion into inference, planning and learning.

## Variational free energy and inference

In variational Bayesian inference (a.k.a., approximate Bayesian inference), model inversion entails the minimisation of variational free energy with respect to the sufficient statistics of approximate posterior beliefs. This can be expressed as follows, where, for clarity, we will deal with a single factor, such that the policy (i.e., combination of paths) becomes the path, $\pi = u$. Omitting dependencies on previous states, we have for model $m$:

$$
\begin{aligned}
Q(s_\tau, u_\tau, a) &= \arg\min_Q F \\
F &= \mathbb{E}_Q[\ln \underbrace{Q(s_\tau, u_\tau, a)}_{posterior} - \ln \underbrace{P(o_\tau \mid s_\tau, u_\tau, a)}_{likelihood} - \ln \underbrace{P(s_\tau, u_\tau, a)}_{prior}] \\
&= \underbrace{D_{KL}[Q(s_\tau, u_\tau, a) \parallel P(s_\tau, u_\tau, a \mid o_\tau)]}_{divergence} - \ln \underbrace{P(o_\tau)}_{evidence} \\
&= \underbrace{D_{KL}[Q(s_\tau, u_\tau, a) \parallel P(s_\tau, u_\tau, a)]}_{complexity} - \underbrace{\mathbb{E}_Q[\ln P(o_\tau \mid s_\tau, u_\tau, a)]}_{accuracy}
\end{aligned}
\tag{1}
$$

Because the (KL) divergences cannot be less than zero, the penultimate equality means that free energy is zero when the (approximate) posterior is the true posterior. At this point, the free energy becomes the negative log evidence for the generative model (Beal, 2003). This means minimising free energy is equivalent to maximising model evidence.

Planning emerges under active inference by placing priors over (controllable) paths to minimise expected free energy (Friston et al., 2015):

$$G(u) = \mathbb{E}_{Q_u}[\ln Q(s_{\tau+1}, a \mid u) - \ln Q(s_{\tau+1}, a \mid o_{\tau+1}, u) - \ln P(o_{\tau+1} \mid c)]$$

$$= -\underbrace{\mathbb{E}_{Q_u}[\ln Q(a \mid s_{\tau+1}, o_{\tau+1}, u) - \ln Q(a \mid s_{\tau+1}, u)]}_{\text{expected information gain (learning)}} -$$

$$\underbrace{\mathbb{E}_{Q_u}[\ln Q(s_{\tau+1} \mid o_{\tau+1}, u) - \ln Q(s_{\tau+1} \mid u)]}_{\text{expected information gain (inference)}} - \underbrace{\mathbb{E}_{Q_u}[\ln P(o_{\tau+1} \mid c)]}_{\text{expected cost}}$$

$$= -\underbrace{\mathbb{E}_{Q_u}[D_{KL}[Q(a \mid s_{\tau+1}, o_{\tau+1}, u) \,\|\, Q(a \mid s_{\tau+1}, u)]]}_{\text{novelty}} + \qquad (2)$$

$$\underbrace{D_{KL}[Q(o_{\tau+1} \mid u) \,\|\, P(o_{\tau+1} \mid c)]}_{\text{risk}} - \underbrace{\mathbb{E}_{Q_u}[\ln Q(o_{\tau+1} \mid s_{\tau+1}, u)]}_{\text{ambiguity}}$$

Here, $Q_u = Q(o_{\tau+1}, s_{\tau+1}, a \mid u) = P(o_{\tau+1}, s_{\tau+1}, a \mid u, o_0, \ldots, o_\tau) = P(o_{\tau+1} \mid s_{\tau+1}, a) Q(s_{\tau+1}, a \mid u)$ is the posterior predictive distribution over parameters, hidden states and outcomes at the next time step, under a particular path. Note that the expectation is over *observations in the future* that become random variables, hence, *expected* free energy. This means that preferred outcomes—that subtend expected cost and risk—are prior beliefs, which constrain the implicit planning as inference (Attias, 2003; Botvinick and Toussaint, 2012; Van Dijk and Polani, 2013).

One can also express the prior over the parameters in terms of an expected free energy, where, marginalising over paths:

$$P(a) = \sigma(-G) \qquad (3)$$
$$G(a) = \mathbb{E}_{Q_a}[\ln P(s \mid a) - \ln P(s \mid o, a) - \ln P(o \mid c)]$$

$$= -\underbrace{\mathbb{E}_{Q_a}[\ln P(s \mid o, a) - \ln P(s \mid a)]}_{\text{expected information gain}} - \underbrace{\mathbb{E}_{Q_a}[\ln P(o \mid c)]}_{\text{expected cost}}$$

$$= -\underbrace{\mathbb{E}_{Q_a}[D_{KL}[P(o, s \mid a) \,\|\, P(o \mid a) P(s \mid a)]}_{\text{mutual information}} - \underbrace{\mathbb{E}_{Q_a}[\ln P(o \mid c)]}_{\text{expected cost}}$$

where $Q_a = P(o \mid s, a) P(s \mid a) = P(o, s \mid a)$ is the joint distribution over outcomes and hidden states, encoded by Dirichlet parameters. Note that the Dirichlet parameters encode the mutual information, in the sense that they implicitly encode the joint distribution over outcomes and their hidden causes. When normalising each column of the *a* tensor, we recover the likelihood distribution (as in Figure 1). However, we could normalise over every element, to recover a joint distribution [as in Equation (5) later].

Expected free energy can be regarded as a universal objective function that augments mutual information with expected costs or constraints. Constraints—parameterised by *c*—reflect the fact that we are dealing with open systems with characteristic outcomes, $o$. This can be read as an expression of the constrained maximum entropy principle (Ramstead et al., 2022). Alternatively, it can be read as a constrained principle of maximum mutual information or minimum redundancy (Ay et al., 2008; Barlow, 1961; Linsker, 1990; Olshausen and Field, 1996). In machine learning, this kind of objective function underwrites disentanglement (Higgins et al., 2021; Sanchez et al., 2019), and generally leads to sparse representations (Gros, 2009; Olshausen and Field, 1996; Sakthivadivel, 2022; Tipping, 2001).

There are many special cases of minimising expected free energy. For example, maximising expected information gain maximises (expected) Bayesian surprise (Itti and Baldi, 2009), in accord with the principles of optimal experimental design (Lindley, 1956). This resolution of uncertainty is related to artificial curiosity (Schmidhuber, 1991; Still and Precup, 2012) and speaks to the value of information (Howard, 1966). Expected complexity or risk is the same quantity minimised in risk sensitive or KL control (Klyubin et al., 2005; van den Broek et al., 2010), and underpins (free energy) formulations of bounded rationality based on complexity costs (Braun et al., 2011; Ortega and Braun, 2013) and related schemes in machine learning; e.g., Bayesian reinforcement learning (Ghavamzadeh et al., 2016). Finally, minimising expected cost subsumes Bayesian decision theory (Berger, 2011).

## Active inference

In variational inference and learning, sufficient statistics—encoding posterior expectations—are updated to minimise variational free energy. Figure 2 illustrates these updates in the form of variational message passing (Dauwels, 2007; Friston et al., 2017a; Winn and Bishop, 2005). For example, expectations about hidden states are a softmax function of messages that are linear combinations of other expectations and observations.

$$\mathbf{s}_\tau^f = \sigma(\mu_{\uparrow\mathbf{A}}^f + \mu_{\to B}^f + \mu_{\leftarrow B}^f + \ldots)$$
$$\mu_{\uparrow\mathbf{A}}^f = \sum_{g \in ch(f)} \mu_{\uparrow\mathbf{A}}^{g,f} \qquad\qquad (4)$$
$$\mu_{\uparrow\mathbf{A}}^{g,f} = \mathbf{o}_\tau^g \odot \varphi(\mathbf{a}^g) \odot_{i \in pa(g) \backslash f} \mathbf{s}_\tau^i$$

Here, the ascending messages from the likelihood factor are a linear mixture[3] of expected states and observations, weighted by (digamma) functions of the Dirichlet counts that correspond to the parameters of the likelihood model (c.f., connection weights). The expressions in Figure 2 are effectively the fixed points (i.e., minima) of variational free energy. This means that message passing corresponds to a fixed-point iteration scheme that inherits the same convergence proofs of coordinate descent (Beal, 2003; Dauwels, 2007; Winn and Bishop, 2005).[4]

---

[3] The notation implies a sum product operator; i.e., the dot or inner product that sums over one dimension of a numeric array or tensor. In this paper, these sum product operators are applied to a vector $\mathbf{a}$ and a tensor $\mathbf{A}$ where, $\mathbf{a} \odot \mathbf{A}$ implies the sum of products is taken over the leading dimension, while $\mathbf{A} \odot \mathbf{a}$ implies the sum is taken over a trailing dimension. For example, $\mathbf{1} \odot \mathbf{A}$ is the sum over rows and $\mathbf{A} \odot \mathbf{1}$ is the sum over columns, where $\mathbf{1}$ is a vector of ones and $\mathbf{A}$ is a matrix. This notation replaces the Einstein summation notation to avoid visual clutter.

## Message passing on factor graphs

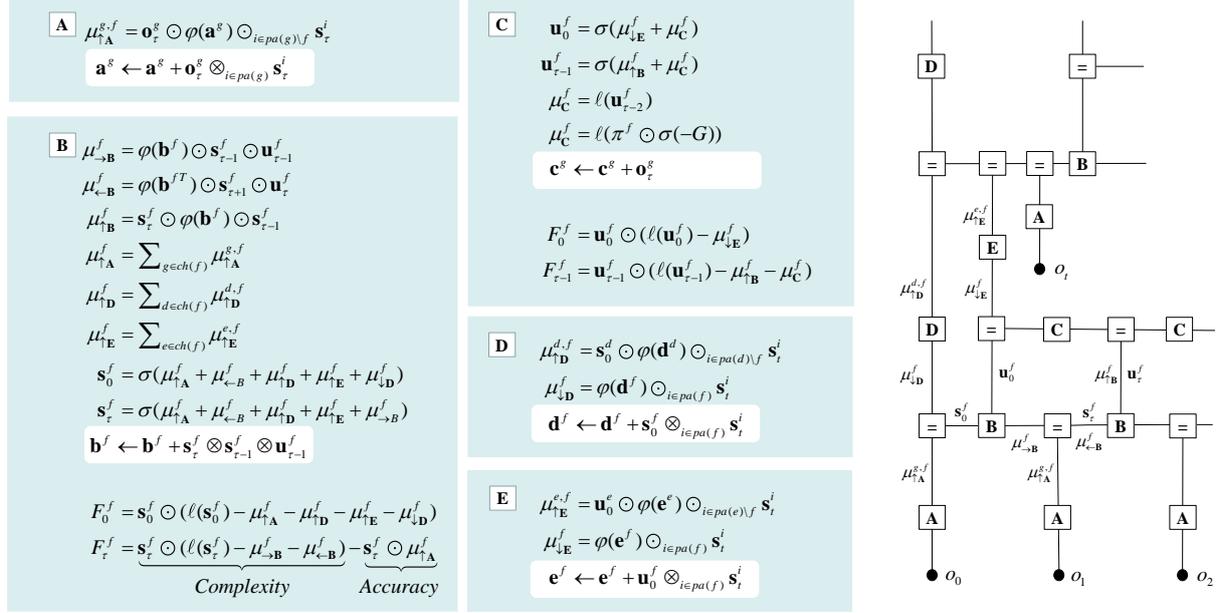

**Figure 2**: **Belief updating and variational message passing**: the right panel presents the generative model as a factor graph, where the nodes (square boxes) correspond to the factors of the generative model (labelled with the associated tensors). The edges connect factors that share dependencies on random variables. The leaves of (filled circles) correspond to known variables, such as observations (*o*). This representation is useful because it scaffolds the message passing—over the edges of the factor graph—that underwrite inference and planning. The functional forms of these messages are shown in the left-hand panels. For example, the expected path—in the first equality of panel **C**—is a softmax function of two messages. The first is a descending message $\mu_{\uparrow E}^f$ from **E** that inherits from expectations about hidden states at the level above. The second is the log-likelihood of the path based upon expected free energy, *G*. This message depends upon Dirichlet counts scoring preferred outcomes—i.e., prior constraints on modality *g*—encoded in $\mathbf{c}^g$: see Figure 1 and Equation (2). The two expressions for $\mu_C^f$ correspond to fixed and control paths, respectively. The updates in the lighter panels correspond to learning; i.e., updating Bayesian beliefs about parameters. Similar functional forms for the remaining messages can be derived by direct calculation. The $\odot$ notation implies a generalised inner product or tensor contraction, while $\otimes$ denotes an outer product. $ch(\cdot)$ and $pa(\cdot)$ return the children and parents of latent variables.

---

## Active learning

Active learning has a specific meaning in this paper. It implies that the updating of Dirichlet counts depends on expected free energy; namely, the mutual information encoded by the tensors: see Equation (1). This means that an update is selected in proportion to the expected information gain. Consider two actions: to update or not to update. From Figure 2, we have (dropping the modality superscript for clarity):

$$\Delta \mathbf{a} = \mathbf{o}_\tau \otimes_{i \in pa} \mathbf{s}_\tau^i \qquad (5)$$

$$\mathbb{E}[P(o, s \mid a)] = \overline{a} =: \frac{a}{\Sigma(a)}$$

$$\mathbb{E}_Q[a \mid u = u_o] = \mathbf{a} \mid u_o = \mathbf{a}$$

$$\mathbb{E}_Q[a \mid u = u_1] = \mathbf{a} \mid u_1 = \mathbf{a} + \Delta \mathbf{a}$$

Here, $\Sigma(a) =: \mathbf{1} \odot a \odot_{i \in pa} \mathbf{1}$ is the sum of all tensor elements. The prior probability of committing to an update is given by the expected free energy of the respective Dirichlet parameters, which scores the expected information gain (i.e., mutual information) and cost[5]:

$$P(u) = \sigma(-\alpha \cdot G(\mathbf{a} \mid u)) \qquad (6)$$

$$G(\mathbf{a}) = -\underbrace{\mathbb{E}_{Q_a}[D_{KL}[Q(s, o \mid \mathbf{a}) \| Q(s \mid \mathbf{a})P(o \mid \mathbf{a})]}_{\text{mutual information}} - \underbrace{\mathbb{E}_{Q_a}[\ln P(o \mid c)]}_{\text{expected cost}}$$

$$= (\mathbf{1} \odot \overline{\mathbf{a}}) \odot \ell(\mathbf{1} \odot \overline{\mathbf{a}}) + (\overline{\mathbf{a}} \odot \mathbf{1}) \odot \ell(\overline{\mathbf{a}} \odot \mathbf{1}) - \mathbf{1} \odot (\overline{\mathbf{a}} \times \ell(\overline{\mathbf{a}})) \odot \mathbf{1} - \varphi(\mathbf{c}) \odot (\overline{\mathbf{a}} \odot \mathbf{1})$$

This prior over the updates furnishes a Bayesian model average of the likelihood parameters, effectively marginalising over update policies:

$$\mathbb{E}_Q[a_{\tau+1}^g] = P(u_0) \cdot \mathbb{E}_Q[a_\tau^g \mid u = u_o] + P(u_1)\mathbb{E}_Q[a_\tau^g \mid u = u_i] \qquad (7)$$

$$\Rightarrow$$

$$\mathbf{a}_{\tau+1}^g = P(u_0) \cdot \mathbf{a}_\tau^g + P(u_1)(\mathbf{a}_\tau^g + \Delta \mathbf{a}_\tau^g)$$

$$= \mathbf{a}_\tau^g + P(u_1)\Delta \mathbf{a}_\tau^g$$

---

[5] For simplicity, we assume tensors have been formatted as matrices, by vectorising the second and trailing dimensions.

In equation (6), $\alpha$ plays the role of a hyperprior that determines the sensitivity to expected free energy. When this precision parameter is large, the Bayesian model average above becomes Bayesian model selection; i.e., either the update is selected, or it is not.

This kind of active learning rests on treating an update as an action that is licensed if expected free energy decreases. A complementary perspective—on this selective updating—is that it instantiates a kind of Maxwell's Demon; selecting just those updates that maximise (constrained) mutual information. Exactly the same idea can be applied to model selection, leading to active selection.

## Active selection

In contrast to learning—that optimises *posteriors* over parameters—Bayesian model selection or structure learning (Tenenbaum et al., 2011; Tervo et al., 2016; Tomasello, 2016) can be framed as optimising the *priors* over model parameters. On this view, model selection can be implemented efficiently using Bayesian model reduction, which starts with an expressive model and removes redundant parameters. Crucially, Bayesian model reduction can be applied to posterior beliefs after the data have been assimilated. In other words, Bayesian model reduction is a *post hoc* optimisation that refines current beliefs based upon alternative models that may provide potentially simpler explanations (Friston and Penny, 2011).

Bayesian model reduction is a generalisation of ubiquitous procedures in statistics (Savage, 1954). In the present setting, it reduces to something remarkably simple: by applying Bayes rules to parent and reduced models it is straightforward to show that the change in variational free energy can be expressed in terms of posterior Dirichlet counts $\mathbf{a}$, prior counts $a$ and the prior counts that define a reduced model $a$'. Using $\mathrm{B}$ to denote the beta function, we have (Friston et al., 2018):

$$
\begin{aligned}
\Delta F &= \ln P(o \,|\, a) - \ln P(o \,|\, a') \\
&= \ln \mathrm{B}(\mathbf{a}) + \ln \mathrm{B}(a') - \ln \mathrm{B}(a) - \ln \mathrm{B}(\mathbf{a} + a' - a) \\
\mathbf{a}' &= \mathbf{a} + a' - a
\end{aligned}
\tag{8}
$$

Here, **a'** corresponds to the posterior that one would have obtained under the reduced priors. Please see (Friston et al., 2020; Smith et al., 2020), for worked examples in epidemiology and neuroscience, respectively.

The alternative to Bayesian model reduction is the bottom-up growth of models to accommodate new data or content. If one considers the selection of one (parent) model over another (augmented) model as an action, then the difference in expected free energy furnishes a log prior over models that can be combined with the (variational free energy bound on) log marginal likelihoods to score their posterior probability. This can be expressed in terms of a log Bayes factor (i.e., odds ratio) comparing the likelihood of two models, given some observation, $o$:

$$\ln \frac{P(m \mid o)}{P(m' \mid o)} = \ln \frac{P(o \mid m)P(m)}{P(o \mid m')P(m')} = \Delta F + \Delta G \qquad (9)$$

$$\Delta F = \ln P(o \mid m) - \ln P(o \mid m')$$
$$\Delta G = \ln P(m) - \ln P(m') = G(\mathbf{a} \mid m) - G(\mathbf{a}' \mid m')$$

Here, $\mathbf{a}$ and $\mathbf{a'}$ denote the posterior expectations of parameters under a parent $m$, and augmented model $m$', respectively. The difference in expected free energy reflects the information gain in selecting one model over the other, following (6). One can now retain or reject the parent model, depending upon whether the log odds ratio is greater than or less than zero, respectively. Active model selection therefore finds structures with precise or unambiguous likelihood mappings. When assimilating new (e.g., training) data one can simply equip the model with a new latent cause to explain each (unique) observation when, and only when, expected free energy decreases (Friston et al., 2023b). This affords a fast kind of structure learning. Before illustrating the above procedures, we now consider a particular structural form that characterises the generative models, used in the illustrative applications.

## Renormalising generative models

Renormalisability is feasible under the models in Figure 1. This follows because the dynamics constitute a coordinate descent on variational free energy, leading to paths of least action; namely, a path integral of variational free energy (Friston et al., 2023a). However, we also require renormalising transformations of model parameters from one hierarchical level to the next. These scale transformations entail a coarse graining that generally induce a separation of temporal scales; such that the dynamics—here, belief updating—slow down, as one ascends levels or scales. The implicit RG-flow rests upon the inclusion of dynamics in the generative model.

In discrete time, the inclusion of paths means that a succession of states at any given level can be generated by specifying the initial state and successive transitions, encoded by the slice of the transition tensor specified by a path. Crucially, the initial state and path can be generated from the superordinate state, which has its own dynamics and associated path. This structure can be read in a number of ways. It can be regarded as a discrete version of switching dynamical systems (Linderman et al., 2016; Olier et al., 2013), in which the switching variables (i.e., paths) change at a slower timescale than the dynamics or paths at the scale being switched. In the limit of continuous time, the composition of implicit RG operators means that one can model changes or switches in velocity—i.e., acceleration— and, at the next level, changes in acceleration—i.e., jerk, and so on. In continuous state-space models, this reduces to working in generalised coordinates of motion (Friston et al., 2010; Kerr and Graham, 2000).

A more intuitive view—of the latitude afforded by temporal renormalisation—is that successively higher levels encode sequences of sequences and, implicitly, compositions of events or episodes. In other words, at a deep level, one state can generate sequences of sequences of sequences; thereby, destroying the Markovian properties of content generated at the lowest level. It is this deep structure that has been leveraged in applications of active inference under continuous models of song and speech; e.g., (Friston and Kiebel, 2009; Yildiz et al., 2013). We will see the discrete homologues of the ensuing semi-Markovian processes later.

In addition to the renormalisation over time, we also have to consider renormalisation over state-space. The example in Figure 3 illustrates a graphical model in which groups of states at a lower level are generated by a single state at the higher level. In the next section, we will associate states at the lowest level with the value of pixels and successive block transformations (i.e., tessellations) with image compression. An important aspect of these models is that the states at any level never share children in the lower level. This renders latent factors at every level conditionally independent. Conditional independence follows from the fact that the Markov blanket of any given state comprises its parents, children and parents of children; however, its children have no co-parents, rendering hidden factors D-separated, when conditioned upon the initial states (and paths) of the level below. In turn, this has the practical implication that the likelihood mappings that link different levels or scales (i.e., the **D** and **E** tensors in Figure 2) are low dimensional matrices at every level of the hierarchy.

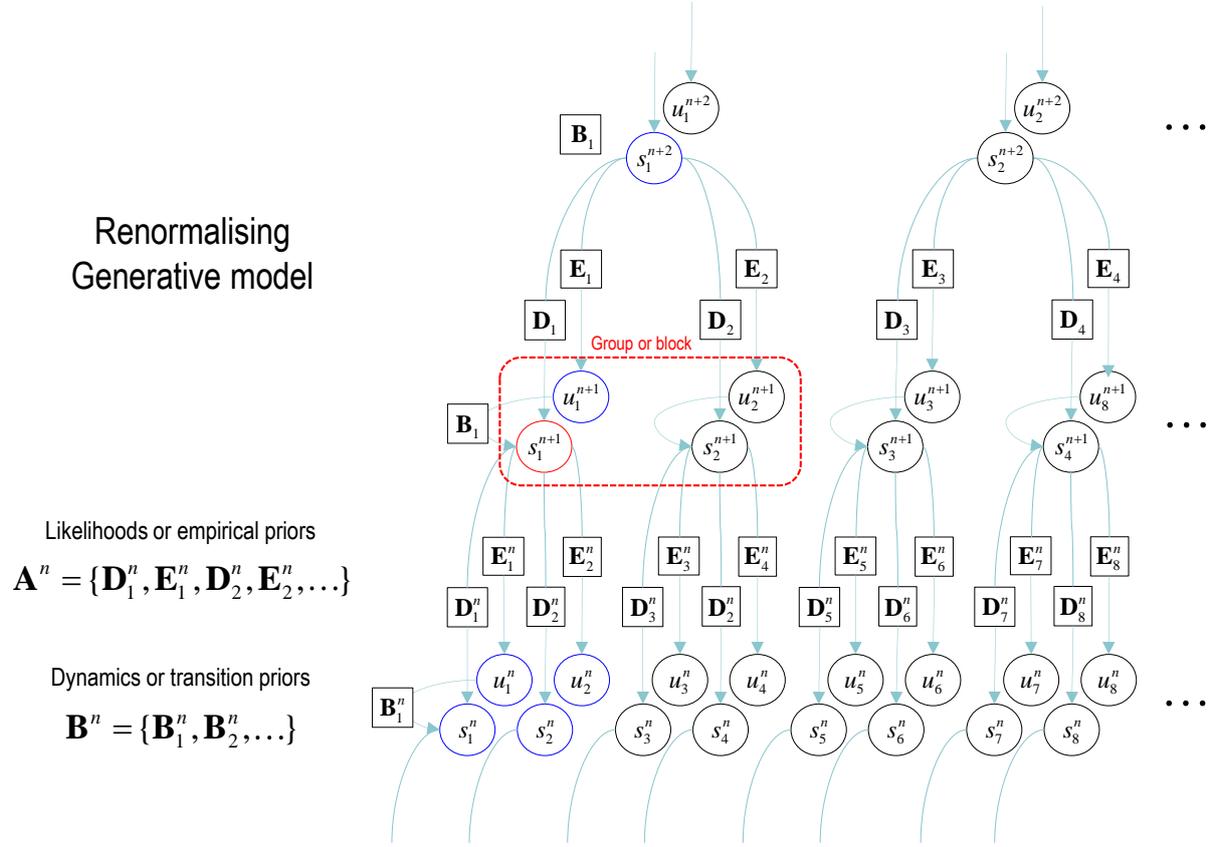

**Figure 3: Renormalising generative model.** This graphical model illustrates the architecture of renormalising generative models (temporal renormalisation has been omitted for clarity). In these models, the latent states at any given level generate the initial conditions and paths of [groups of] states at the lower level (red box). This means that the trajectory over a finite number of timesteps at the lower level is generated by a higher state. This entails a separation of temporal scales and implicit renormalisation, such that higher states only change after a fixed number of lower state transitions. This kind of model can be specified in terms of (i) transition tensors ($\mathbf{B}$) at each level, encoding transitions under each discrete path and (ii) likelihood mappings between levels, corresponding to the $\mathbf{D}$ and $\mathbf{E}$ tensors of previous figures. These can be treated as subtensors of likelihood mappings $\mathbf{A}^n = \{\mathbf{D}_1^n, \mathbf{E}_1^n, \mathbf{D}_2^n, \mathbf{E}_2^n, \ldots\}$ that furnish empirical priors over the states (and paths) at each level ($n$). Because each state (and path) has only one parent, the ensuing Markov blanket (blue circles) of each state (red circle) ensures conditional independence among latent factors. In sum, a renormalising generative model (RGM) is a hypergraph, parameterised by two sorts of ($\mathbf{A}$ and $\mathbf{B}$) tensors, in which the children of states at any level bipartition into initial states and paths. In this example, the blocking transformation groups pairs of states (and paths).

Models of this sort can be regarded as generating local dependencies within each group of states at the lower level, where global (between-group) dependencies are modelled at higher levels. They feature a characteristic progression from local to global as one ascends the hierarchy: c.f., (Hochstein and Ahissar, 2002), providing an efficient way to model data that show strong local dependencies over space and time[6].

In the setting of discrete models, one can also regard an RGM as an expressive way of modelling nonlinearities in the generation of data; much in the spirit of deep neural networks with nonlinear activation or rectification functions. In the limit of precise likelihood mappings, this can be viewed as the composition of logical operators. For example, if the first two elements of the leading dimension of a **D** likelihood tensor were nonzero, this means that the parent state can generate the first OR second child's initial state. Conversely, nonzero elements of two likelihood tensors generate a particular state of one child AND another child. Heuristically, one can see why an RGM can compose operators to represent or generate content that has compositional structure. The inversion of an RGM could be construed as a simple form of abductive (e.g., modal) logic. In the ensuing example, this perspective is leveraged to learn the composition of local image features subtending global objects.

# IMAGE COMPRESSION AND COMPOSITIONALITY

We start with the simplest example of renormalising generative models apt for image classification and recognition. The aim here is to automatically assemble an RGM to classify and generate the image content to which it is exposed. In other words, we seek procedures for automatic structure learning—followed by learning the parameters of the ensuing structure—which can then be deployed to classify or generate the kind of content on which it was trained.

The first step is to quantise continuous pixel values for an image of any given size. Effectively, this involves mapping continuous pixel values to some discrete state space, whose structure can be learned automatically by recursive applications of a blocking transformation. In this example, we group together local pixels by tiling (i.e., tessellating) an image, partitioning it into little squares with a spin-block transformation (Vidal, 2007)[7]. Each group is then subject to singular value decomposition, given a training set of images, to identify an orthogonal (spatial) basis set of singular vectors. This grouping is followed by a reduction operator that by retaining singular variates with large singular values (here, the first 32 principal vectors based upon groups of 4x4 pixels). Under linear transformations, this is guaranteed to maximise the mutual information in accord with Equation (3) (given the absence of prior constraints). One way to see this is to assume a precise conditional distribution of a continuous variable $I$ given observations that can be expressed in terms of linear weights inside a Dirac delta, and identify the weights that maximise the mutual information (using $\rho$ to indicate a mean-centred sample distribution):

$$p(I \mid o) = \delta(o - W \cdot I) \Rightarrow D_{KL}[\rho(I)p(o \mid I) \| \rho(I)\rho(o)]$$
$$= -\mathbb{E}_{\rho(o)}[\ln \rho(o)] \approx \tfrac{1}{2} \ln |(W^T U S)^2|$$
$$\Rightarrow U \propto \arg \max_W |(W^T U S)^2| \tag{10}$$

$$\mathbb{E}_{\rho(o)}[I \odot I^T] = U S^2 U^T, \quad [I_1, I_2, ...] = U S V^T$$

---

[6] A more thorough treatment of the implicit RG operator or blocking transformation would rest on characterising conditional independencies within and between groups, in terms of their Markov blankets Friston, K.J., Fagerholm, E.D., Zarghami, T.S., Parr, T., Hipólito, I., Magrou, L., Razi, A., 2021c. Parcels and particles: Markov blankets in the brain. Network Neuroscience 5, 211-251. However, for systems with strictly local conditional dependencies (e.g., Markov random Fields) the Markov blanket of any element constitutes its nearest neighbours. This motivates the use of the spin-block transformations: e.g., Kaluza, P., Meyer-Ortmanns, H., 2010. On the role of frustration in excitable systems. Chaos 20, 043111, Vidal, G., 2007. Entanglement renormalization. Physical Review Letters 99.

[7] In practice, we use overlapping groups, where the singular value decomposition is applied following weighting by a radial (Gaussian) basis function whose standard deviation is the distance between group centres.

The conditional entropy of the delta distribution tends to zero and is therefore omitted on the second line. The approximate equality rests upon an assumption that the sample distribution for $o$ is approximately Gaussian and makes use of a singular value decomposition of the concatenated samples, where the diagonal elements of $S$ are singular values. To maximise the entropy of the marginal of $o$, the weights must be chosen to be proportional to the left singular vectors (columns of $U$) or, equivalently, the eigenvectors of the sample covariance.

The set of singular variates for each group specifies the pattern for any given image at the corresponding location. The continuous variates then be quantised to a discrete number of levels (here, seven) to provide a discrete representation of each block. This corresponds to the first RG operator (a.k.a., blocking transformation). Given a partition of the image into quantised blocks, we now apply a second block transformation into groups of four nearest neighbours. This reduces the number of blocks by a factor of two in each image dimension. One then repeats this procedure until there is only one group at the highest level or scale.

Each application of the block transformation creates a likelihood mapping (**D**) from the states at a higher level to the lower level. In other words, the state of a latent factor at any level generates the states of a group at lower levels (or quantised singular variates for pixels within a group at the image level). The requisite likelihood matrices can be assembled using a fast form of structure learning based upon Equation (9). This equation says that the likelihood mapping (in the absence of any constraints) should have the maximum mutual information. This is assured if each successive column of the likelihood matrix is unique. In turn, this means we can automatically assemble the requisite likelihood mappings by appending *unique* instances of quantised singular variates (encoded as one-hot vectors) in a set of training images. This results in one-hot likelihood arrays for each group that share the same parent at the higher level:

$$
\begin{aligned}
\mathbf{A}^{\ell} &= \mathbf{D}_1^{\ell} \\
\mathbf{A}^{\ell-1} &= \{\overbrace{\mathbf{D}_1^{\ell-1}, \mathbf{D}_2^{\ell-1}, \mathbf{D}_3^{\ell-1}, \mathbf{D}_4^{\ell-1}}^{pa(\mathbf{D}^{\ell-1})=1}\} \\
&\vdots \\
\mathbf{A}^{n+1} &= \{\overbrace{\mathbf{D}_1^{n+1}, \mathbf{D}_2^{n+1}, \mathbf{D}_3^{n+1}, \mathbf{D}_4^{n+1}}^{pa(\mathbf{D}^{n+1})=1}, \overbrace{\mathbf{D}_5^{n+1}, \mathbf{D}_6^{n+1}, \mathbf{D}_7^{n+1}, \mathbf{D}_8^{n+1}}^{pa(\mathbf{D}^{n+1})=2}, \ldots\} \\
\mathbf{A}^{n} &= \{\overbrace{\mathbf{D}_1^{n}, \mathbf{D}_2^{n}, \mathbf{D}_3^{n}, \mathbf{D}_4^{n}}^{pa(\mathbf{D}^{n})=1}, \overbrace{\mathbf{D}_5^{n}, \mathbf{D}_6^{n}, \mathbf{D}_7^{n}, \mathbf{D}_8^{n}}^{pa(\mathbf{D}^{n})=2}, \overbrace{\mathbf{D}_9^{n}, \mathbf{D}_{10}^{n}, \mathbf{D}_{11}^{n}, \mathbf{D}_{12}^{n}}^{pa(\mathbf{D}^{n})=3}, \ldots\} \\
&\vdots
\end{aligned}
\tag{11}
$$

We have dropped $\mathbf{E}^n = [1,1,\ldots]$ because there is only one path in the absence of dynamics. In this case, the penultimate level comprises the likelihood mappings that generate each quadrant of the image in terms of quadrants of quadrants, much like a discrete wavelet decomposition. The ultimate level corresponds to priors over the (initial) states or class of image. In short, structure learning emerges from the recursive application of blocking transformations of some training images. This is a special case of fast structure learning (see Algorithm 1), described in detail in the next section

At first glance, this procedure may appear to generate likelihood mappings (**D** matrices) of increasing size, because the combinatorics of increasingly large groups could explode at higher levels. However, by training on a small number of images, one upper bounds the number of states at each level. This follows from the fact that there can be no more columns of the likelihood matrix than there are unique images in the training set. In other words, one can effectively encode a finite number of images— without any loss of information—such that RGM inversion corresponds to lossless compression.

To generalise to a more expansive training set, one can populate the likelihood mappings with small concentration parameters and use active learning to recover an optimal lossy compression. Active learning—according to Equation (7)—simply means accumulating appropriate Dirichlet counts in the likelihood mappings until the mutual information converges to its maximum. This offers a principled way to terminate the ingestion of training data; after which, there can be no further improvement in expected free energy or mutual information.

## A worked example

To demonstrate the above methods, they were applied to the MNIST digit classification problem (LeCun and Cortes, 2005). MNIST images were pre-processed by up-sampling to 32×32 pixels, smoothing and histogram equalisation[8]. In addition, they were converted into a format suitable for video processing with three (TrueColor, RGB) channels. An exemplar image is shown in Figure 3 (left panel). The blocking transformation to discrete state-space is illustrated in the right panel, which shows the reconstructed image (i.e., local mixtures of singular vectors weighted by discrete singular variates). The centroids of each group are shown with small red dots, where each group comprised pixels within a radius of four pixels.

Based upon the prior that there can be a dozen ways of writing any given number, the first 13 (Baker's dozen) images of each digit class were used for fast structure learning. This produced an RGM with four levels. The centroids of the ensuing groups—of increasing size—are shown as the successively larger red dots in Figure 4. At the penultimate level, this grouping is into quadrants. In this application we equipped the last level—covering all pixel locations—with a likelihood mapping between the known digit class ( i.e., label) and compressed representations at the penultimate level.

The likelihood mappings are shown in Figure 5 to illustrate the ensuing structure. The lower  row shows concatenated likelihood mappings at each level. The upper row reproduces these mappings after transposing, to illustrate how states at one level generate the states of groups at the subordinate level. For example, at level 1, we have 16 groups of pixels, whose states are generated by four groups at level 2. Similarly, the four groups at level 2 are generated by one group at a level 3. Level 4 implements our prior knowledge about digit classes, effectively providing a mapping from digit class to the (13×10) exemplar images that have been compressed in a lossless fashion.

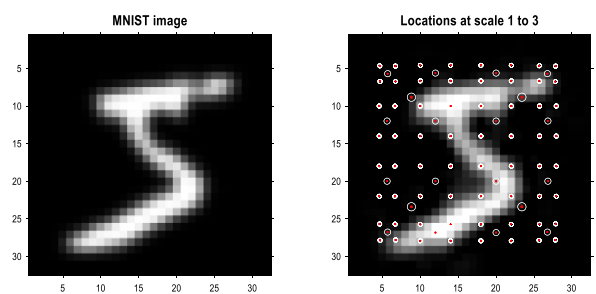

**Figure 4**: **quantising images**. The left panel shows an example of an MNIST image after resizing to 32×32 pixels, following histogram equalisation. The image in the right panel corresponds to the image in pixel space generated by quantised singular variates, used to form a linear mixture of singular vectors over groups of pixels. The centres of the (4x4) groups of pixels are indicated by the small red dots (encircled in white). In this example, the singular variates could take seven discrete values—centred on zero—for a maximum of 16 singular vectors. At subsequent

---

[8] The digits were downloaded from https://lucidar.me/en/matlab/load-mnist-database-of-handwritten-digits-in-matlab/ and pre-processed by smoothing with Gaussian convolution kernel (of two pixels width).

levels, (2×2) groups of groups are combined via grouping or blocking operators. The centroids of these groups (of groups) at the three successive scales are shown with successively larger red dots. At the third scale, there are four groups corresponding to the quadrants of the original image.

$$\mathbf{A}^n = \{\mathbf{D}_1^n, \mathbf{E}_1^n, \mathbf{D}_2^n, \mathbf{E}_2^n, \ldots\}$$

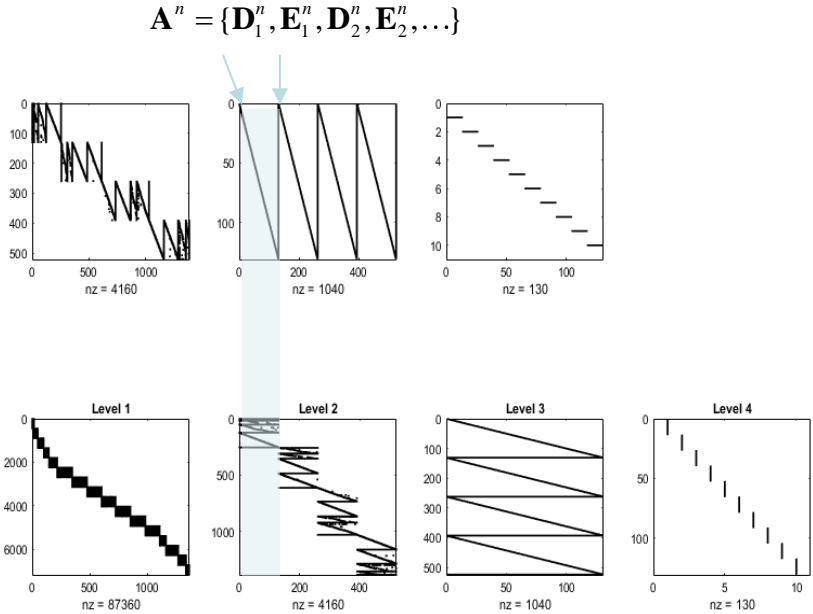

**Figure 5: Renormalising likelihoods**. This figure is a schematic representation of the composite likelihood mappings (comprising **D** and **E**) among the levels of an RGM, following fast structure learning. By the third level, each latent state can generate an entire image via recursive application of sparse, block diagonal matrices, where 'nz' counts the number of non-zero elements. In this example, the model has been equipped with a fourth level, mapping from 10 digit classes to 130 latent states at the third level (encoding 13 images of 10 digits). These likelihood mappings (that mediate empirical priors) are assembled automatically during structure learning by accumulating unique combinations of (recursively grouped) states at subordinate levels. The upper row reproduces the matrices of the lower row—after transposition—to illustrate the dimension reduction inherent in the grouping of states. For example, the thousand or so states at the second level generate over 6000 states at the first, which specify the mixture of singular vectors required to generate an image. Similarly, the 500 or so states at level 3 generate empirical priors over a partition of level 2 states into four subsets or groups (the first is highlighted in cyan), and so on. Crucially, by construction, the children of states at any level constitute a partition, such that every child is included in exactly one subset. This means that states at any level have only one parent, rendering the subsets of the partition at the higher level conditionally independent. In other words, there are no conditionally dependent co-parents. This enables efficient sum-product operations, during model inversion, because one only has to compute dot products of subtensors (i.e., small matrices) specified by the parents of a group. Note that the matrices in this figure are not simple likelihood mappings: they are concatenated likelihood mappings from all hidden states at one level to all hidden states (and paths) at the subordinate level, where the sum to one constraint is applied to the states (or paths) each child could be in.

Note that the graphics in Figure 5 represent concatenated likelihood matrices. In other words, each block of the likelihood matrices generates multiple outcomes; namely, combinations of outputs for several groups. This illustrative concatenation conceals the fact that each likelihood (i.e., **A**, **D** or **E** tensor) is a relatively small matrix. The implicit sparsity of these likelihoods inherits from the blocking transformations based upon a partition at each level. If we had imposed the additional constraint (i.e., structural prior) that these likelihood mappings are conserved identically over groups, then one would have the discrete homologue of a convolutional neural network, with an implicit weight sharing. Here, we did not impose this prior constraint because different groups of pixels show systematic, location-dependent differences. From a biomimetic perspective, this can be likened to differences in the size of receptive fields between central (i.e., foveal) and peripheral visual fields that speaks to the principles of maximum mutual information and minimum redundancy (Barlow, 1961; Linsker, 1990; Olshausen and Field, 1996; Simoncelli and Olshausen, 2001). Technically, this is reflected in the fact that the peripheral groups of pixels showed little or no variation over exemplar images. This means that only a small number of singular variates were retained in the periphery—explaining why the size of the groups at the first level increases towards the centre of the image (see the lower left panel of Figure 5).

Pursuing the biomimetic theme, Figure 6 illustrates the encoding of images at successively deeper levels of the RGM. The top row (Scale 4) shows the images generated under the first eight levels of the corresponding latent factor. By construction, this factor has 10 levels, corresponding to prior knowledge about the class of each digit. The subsequent rows illustrate the projective fields of particular states at lower levels. These fields can be regarded as the changes in predicted stimuli due to changes in the representations (i.e., the complement of receptive fields). In Figure 6, the first state of every factor of every level was selected to provide a baseline image. This was then subtracted from the image generated by subsequent states of the first factor. The purpose of these characterisations was to show that changing posterior expectations at the highest levels (Scale 3 and 4) produces changes everywhere in the image. In other words, these are coarse-grained global representations of the *object* represented. Conversely, at subordinate levels (Scale 1 and 2) the projective fields are restricted to local parts of the image. The form of these projective fields is remarkably similar to that seen in the visual system; namely, compact simple receptive fields in early visual cortex and complex fields (e.g., with a centre-surround structure) of greater spatial extent at higher levels in the visual hierarchy (Angelucci and Bullier, 2003; Zeki and Shipp, 1988). It is noteworthy that this kind of functional specialisation is an emergent property of a renormalising generative model.

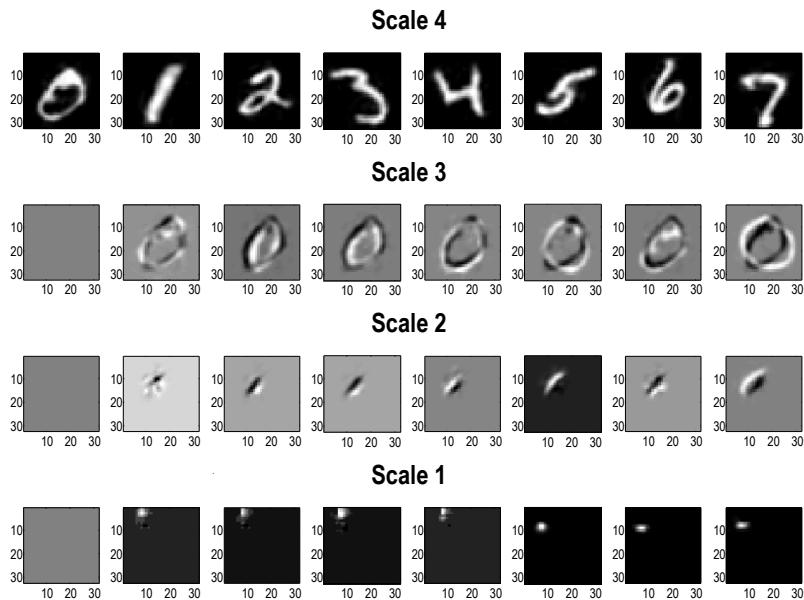

**Figure 6: Renormalisation and projective fields**. This figure shows exemplar projective fields of the (MNIST) RGM in terms of posterior predictive densities in pixel space associated with states at successive levels (or scales) in the generative model. The top row corresponds to the posterior predictions of the first eight states at the fourth level, while subsequent rows show the differences in posterior predictions obtained by switching the first state for the subsequent eight states at each level. The key thing to note is that the size of the projective fields become progressively smaller and more localised, as we descend scales or levels.

From a statistical or Gestalt perspective, the progressive enlargement of projective fields can also be viewed in terms of composition and the transformation from local—place coded—representations to global—object centred—representations: e.g., (Hochstein and Ahissar, 2002; Kersten et al., 2004). This is again reminiscent of biomimetic architectures in the sense that elemental image features are encoded at the lower levels of a visual hierarchy while objects and classes (e.g. faces) are encoded at much higher levels with successive levels composing more extended but compressed representations from lower level representations (Alpers and Gerdes, 2007; Kersten et al., 2004; Zeki and Shipp, 1988).

The foregoing illustrates the generation of content following structure learning of a lossless sort. We now turn to inference and classification. This rests upon optimising the parameters of the RGM structure to maximise the marginal likelihood of some training data. Figure 5 illustrates the results of this active learning during exposure to the first 10,000 training images of the MNIST dataset. This training proceeded by (i) populating all the likelihood mappings with small concentration parameters[9] and (ii) equipping the highest level of the model with precise priors ($\mathbf{D}^4$), corresponding to the class labels and

---

[9] Adding a concentration parameter of 1/16 to the one-hot likelihood arrays produced by fast structure learning and renormalising.

(iii) accumulating Dirichlet parameters according to Equation (7), with $\alpha = 512$ .

This kind of evidence accumulation converges when the expected free energy asymptotes. This convergence is illustrated in the middle panel of Figure 7, which plots the mutual information at the final level as a function of training exemplars. The broken red line corresponds to the upper bound on mutual information afforded by the fact that there are 10 classes or hidden states at the final level. The left panel shows the equivalent (summed) mutual information of lower level likelihood mappings, while the right panel shows the accompanying variational free energy following each exemplar. One can see that there is an initial period of fast learning that asymptotes in terms of mutual information and (negative) variational free energy with little further improvement after about 5,000 samples. Note that the expected free energy provides a convergence criterion that quantifies the number of training exemplars required to maximise model evidence (on average).

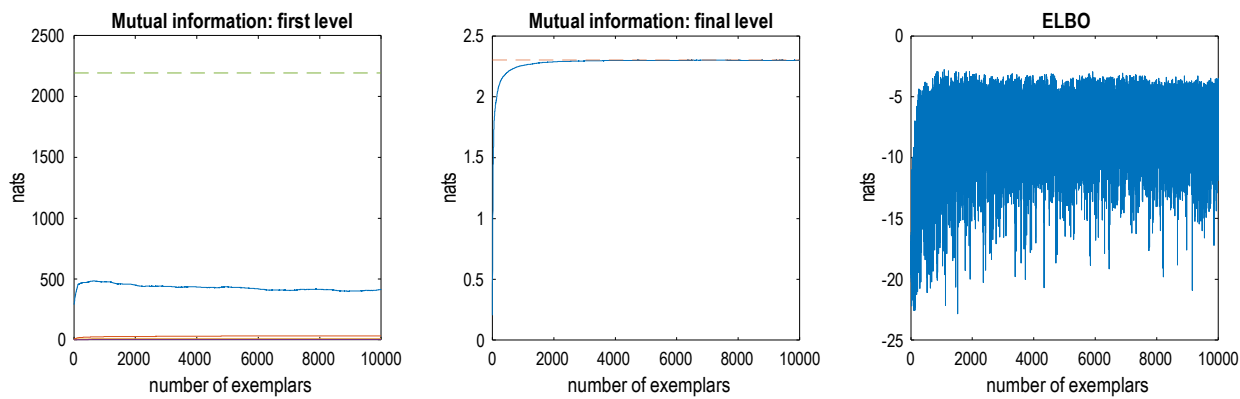

**Figure 7: Active learning**. This figure reports the assimilation or active learning of the training MNIST dataset. In this example, images were assimilated if, and only if, they increase the expected free energy of the RGM. In the absence of prior preferences or constraints, this ensures minimum information loss by underwriting informative likelihood mappings at each level (i.e., maximising mutual information). The left panels report the mutual information as a function of ingesting 10,000 training images. The left panel reports the mutual information at the first level (blue line) and intermediate levels. The middle panel reports the mutual information at the final (fourth) level. The dashed lines correspond to the maximum mutual information that could be encoded by the likelihood mappings. The right panel shows the corresponding evidence lower bound (negative variational free energy), scored by inferring the latent states (digit class) generating each image. The fluctuations here reflect the fact that some images are more easily explained than others, under this model. As the model improves, there are progressively fewer images with a very low ELBO: i.e., –16 natural units or less.

Figure 8 reports the capacity of the ensuing model to correctly infer or classify the class of each the subsequent 1000 (unseen test) images. This figure provides a nuanced report of classification performance, which reflects the fact that we have two kinds of inference at hand. These rest on (i) the posterior distribution over digit classes and (ii) the marginal likelihood that the image is classifiable under the model. This means that one can assess the accuracy of classification conditioned upon whether a particular image was classifiable; enabling a more comprehensive characterisation of performance in terms of sensitivity and specificity. The format adopted in Figure 8 plots the classification accuracy as a function of the ELBO (right panel) and the distribution of ELBOs for correctly and incorrectly classified images (left panel). The thing to note here is that classification accuracy increases with the marginal likelihood that each image belongs to the class of numbers. For example, if we split the test data into two halves, the half with the highest marginal likelihood was classified with 99.8% accuracy, while the classification accuracy for the entire test set was only 95.1%

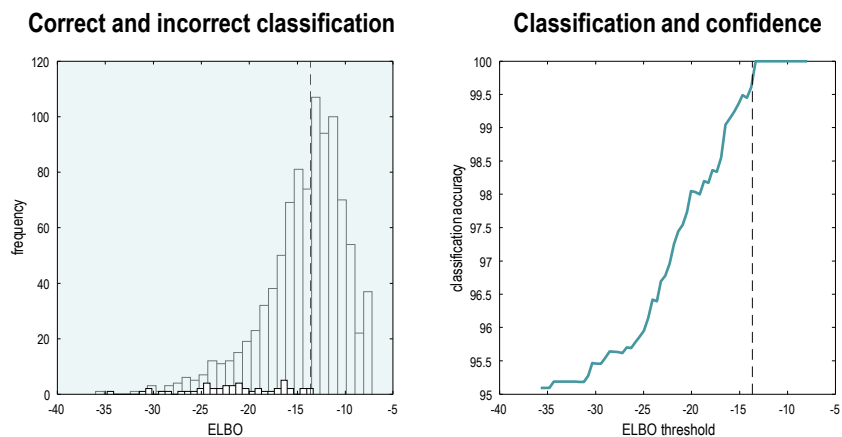

**Figure 8: Classification and confidence**. This figure reports classification performance following the learning described Figure 7. Because inverting a generative model corresponds to inference, recognition or classification, one can evaluate the posterior over latent causes—here, digit class—and the marginal likelihood (i.e., model evidence) of an image, while accommodating uncertainty about its class. This means that one can score the probability that each image was caused by any digit class in terms of the ELBO. The distribution of the ELBO over the 10,000 training images is shown as a histogram in the left panel (for correctly classified images). The smaller histogram (foregrounded) shows the distribution of log-likelihoods for the subset of images that were classified incorrectly. Having access to the marginal likelihood means that one can express classification accuracy as a function of the (marginal) likelihood the image was generated by a digit. The ensuing classification accuracy is shown in the right panel as a function of a threshold (c.f., Occam's window) on the ELBO or evidence that each image was generated by a digit. The vertical dashed lines show the median ELBO (-13.85 nats). Classification accuracy for all images was only 95.1%. However, the accuracy rises to 99.8% following a median split based on their marginal likelihoods.

Images with a low marginal likelihood can be regarded as ambiguous or difficult to classify because they have a small likelihood of being sampled from the class of digits. Figure 9 provides some examples, which speak to the potential importance of scoring the validity of classification (Bach et al., 2015).

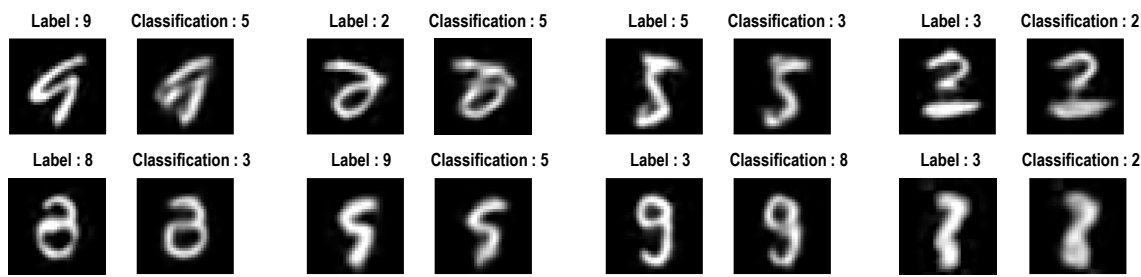

**Figure 9: classification failures**. This figure provides examples of incorrect classification of images with a small marginal likelihood. Each pair of images presents the training image with its label and the corresponding posterior prediction in pixel space and accompanying maximum *a posteriori* classification.

## Summary

This section illustrates the use of renormalisation procedures for learning the structure of a generative model for object recognition—and generation—in pixel space. The protocol uses a small number of exemplar images to learn a renormalising structure apt for lossless compression. The ensuing structure was then generalised by active learning; i.e., learning the likelihood mappings that parameterise the block transformations required to compress images sampled from a larger cohort. This active learning ensures a high mutual information between the scale-invariant mapping from pixels to objects or digit classes. Finally, the RGM was used to classify test images by inferring the most likely digit class.

It is interesting to compare this approach to learning and recognition with the complementary schemes in machine learning. First, the supervision in active inference rests on supplying a generative model with prior beliefs about the causes of content. This contrasts with the use of class labels in some objective function for learning. In active inference, the objective function is a variational bound on the log evidence or marginal likelihood. Committing to this kind of (universal) objective function enables one to infer the most likely cause (e.g., digit class) of any content and whether it was generated by any cause (e.g., digit class), *per se*.

In classification problems of this sort, test accuracy is generally used to score how well a generative model or classification scheme performs. This is similar to the use of cross-validation accuracy based upon a predictive posterior. The key intuition here is that test and cross-validation accuracy can be read as proxies for model evidence (MacKay, 2003). This follows because log evidence corresponds to accuracy minus complexity: see Equation (2). However, when we apply the posterior predictive density to evaluate the expected log likelihood of test data, the complexity term vanishes, because there is no further updating of model parameters. This means, on average, the log evidence and test or cross-validation accuracy are equivalent (provided the training and test data are sampled from the same distribution). Turning this on its head, models with the highest evidence generalise, in the sense that they furnish the highest predictive validity or cross validation (i.e., test) accuracy. One might argue that the only difference between variational procedures and conventional machine learning is that variational procedures evaluate the ELBO explicitly (under the assumed functional form for the posteriors), whereas generic machine learning uses a series of devices to preclude overfitting; e.g., regularisation, mini-batching, and other stochastic schemes. See (Sengupta and Friston, 2018) for further discussion.

This speaks to the sample efficiency of variational approaches that elude batching and stochastic procedures. For example, the variational procedures above attained state-of-the-art classification accuracy on a self-selected subset of test data after seeing 10,000 training images. Each training image was seen once, with *continual learning* (and no notion of batching). Furthermore, the number of training images actually used for learning was substantially smaller[10] than 10,000; because active learning admits only those informative images that reduce expected free energy. This (Maxwell's Demon) aspect of selecting the right kind of data for learning will be a recurrent theme in subsequent sections.

Finally, the requisite generative model was self-specifying, given some exemplar data. In other words, the hierarchical depth and size of the requisite tensors were learned automatically within a few seconds on a personal computer. In the next section, we pursue the notion of efficiency and compression in the context of timeseries and state-space generative models that are renormalised over time.

# Video compression and generative AI

This section generalises the renormalising procedures of the previous section to include dynamics for recognising and generating ordered sequences of images. Procedurally, this simply involves specifying a scale transformation in time, and installing unique state transitions into the prior transition tensors at each level of the RGM. In this setting, structure learning reduces to quantising images in space, colour and time to produce time-colour-pixel *voxels*. Unique transitions among neighbouring voxels are then recorded in **B** tensors; where each unique voxel state is recorded in the column the corresponding **A** matrix of each group of voxels. The timeseries of quantised images is then partitioned into segments of equal lengths (e.g., pairs) and the ensuing segments are inverted (using the **A** and **B** tensors) to evaluate posterior estimates over the initial state and path, of each segment[11]. This produces a new sequence that is coarse-grained in time. Neighbouring states (and paths) are then grouped together and the process is repeated to populate the likelihood mappings (**D** and **E**) from parents at the higher level. By repeating this process, one ends up with a representation of an image sequence in terms of paths through the states of a single factor at the highest level. Each state in this factor generates the initial state (and path) of a group at the lower level, and so on recursively, until an image sequence is generated.

---

[10] In these numerical studies, about 20% of training images were accumulated during active learning.

[11] Model inversion is used so that this fast structure learning can be applied to categorical probability distributions over quantised singular variates; e.g., when supplied with event-related averages of quantised images.

Algorithm 1 provides a pseudocode description of the procedure, where $R$ denotes the number of frames constituting a video event and $j$ labels successive blocks in terms of unique instances. Here, the renormalisation—implicit in fast structure learning—is expressed in terms of the Dirichlet parameters of the requisite tensors: $\mathbf{a}^n = \{\mathbf{d}_1^n, \mathbf{e}_1^n, \mathbf{d}_2^n, \mathbf{e}_2^n, \ldots\}$.

---

Algorithm 1 : *fast structure learning*

---

// reshape and partition video frames

$\{I_1, I_2, \ldots\} \leftarrow I \in \mathbb{R}^{pixels \times pixels \times colours \times frames}$

$I_b \in \mathbb{R}^{(block \times colours \times R) \times T} : T = \frac{frames}{R}$

for every $b$

     $[U_b, S_b, V_b] = SVD(I_b)$          // singular value decomposition

     $U_b \leftarrow U_b \odot S_b \in \mathbb{R}$            // scale singular variates

     $\mathbf{o}_b^0 \leftarrow Bin(U_b) \in (0,1)$        // and quantise into bins

end for

// recursive renormalisation

while   $\#(\mathbf{o}^n) > 1$

     $\{\mathbf{o}_1^n; \mathbf{o}_2^n; \ldots\} \leftarrow \mathbf{o}^n$          // partition into spatial blocks

     $\mathbf{o}_b^n = \{\mathbf{o}_{b_g}^n; \ldots\}$               :     $\mathbf{o}_{b_g}^n \in (0,1)$

     for every $b$                // accumulate unique blocks

         $j = unique(\mathbf{o}_b^n) \in \mathbb{N}^{1 \times T}$     // find unique blocks

         for every $\tau$

              $\mathbf{a}_{\bullet, j(\tau)}^{n,g} = \mathbf{o}_{b_g, \tau}^n$      :    $pa(g) = f = j(\tau)$

             $\mathbf{b}_{j(\tau+1), j(\tau), u}^{n,f} = 1,$    s.t.    $\mathbf{b}_{j(\tau+1), j(\tau), i<u}^{n,f} = 0$

         end for

     end for

     $\{\mathbf{o}_1^n, \mathbf{o}_2^n, \ldots\} \leftarrow \mathbf{o}^n$          // partition into temproal blocks

     $\mathbf{o}_\tau^n \in (0,1)$

     for every $\tau$               // accumulate unique events

         $\mathbf{s}_\tau^n = P(s_0 \mid \mathbf{o}_\tau^n, \mathbf{a}^n, \mathbf{b}^n)$    // initial states

         $\mathbf{u}_\tau^n = P(u \mid \mathbf{o}_\tau^n, \mathbf{a}^n, \mathbf{b}^n)$     // and paths

         $\mathbf{o}_\tau^{n+1} \leftarrow (\mathbf{s}_\tau^n; \mathbf{u}_\tau^n)$       // concatenate

     end for

end while

---

By partitioning each sequence into nonoverlapping pairs—during the time-scale transformation at successive levels—one effectively halves the length of the sequence when ascending from one level to the next. From the perspective of generating sequences, this means that a state at one level generates two successive states at the lower level: by generating the initial state and the path to the second state.

The ensuing RGM acquires a definitive feature of such models; namely, a separation of temporal scales, in which there are two (or more) belief updates for every update at the level above. This means that higher levels encode sequences of sequences of sequences that can be regarded as *episodes* of successive *events*. From the perspective of recognition or classification, an image sequence is successively compressed—with a coarse graining or blocking over space and time—into a succession of events. This furnishes an *event-based* representation that generalises the *object-based* representations of the preceding section. Note that this kind of RGM compresses images or scenes that may involve multiple objects or, indeed, causes that may not have the attribute of objecthood; such as textures and backgrounds.

During inference, the separation of temporal scales manifests a particular kind of reactive message passing (Bagaev and de Vries, 2021; Hewitt et al., 1973). Because each level generates a small sequence (here pairs), the variational message passing depicted in Figure 2 is scheduled as follows: at the highest level, priors are passed to the level below to generate two iterations. After two belief updates, the ascending messages are returned to the higher level to form a posterior over events. However, before the lower level can respond, it has to query its lower level, waiting for two iterations before updating its beliefs, and so on, down to the first level. This means any given level receives messages or requests from the level above—and responds to those requests—at a slower rate than it exchanges messages with its subordinates. In short, lower levels update their beliefs more quickly, in a way that rests upon an asymmetry in the frequency of hierarchical message passing—an asymmetry that characterises message passing in the real neuronal networks (Bastos et al., 2015; Hasson et al., 2008; Kiebel et al., 2008; Pefkou et al., 2017).

## A worked example

To illustrate the basic architecture of this RGM, we used a short video sequence of a Dove flapping her wings. The original video[12] was down-sampled to 32 frames, where each frame comprised 128×128 pixels with three TrueColor channels. The first scaling transformation—from images to discrete state-space—used the same nearest-neighbour block transformation as in the previous section; however, here, we blocked each image into 4x4 blocks, generating an RGM with two hierarchical levels. Crucially, singular value decomposition operated on successive pairs of images (c.f., spatiotemporal receptive fields). This simply involved reshaping the tensors for each image block to concatenate colour and time before applying a singular value decomposition. The ensuing singular vectors therefore span three colours and two time points, compressing the video into 64 time-colour-pixel *voxels*. The remaining scaling transformations constructed the requisite likelihood and transition matrices as described in Algorithm 1.

---

[12] From https://pixabay.com/videos/search/birds/

Figure 10 shows the first frame from the original video and the reconstructed frame following compression using the same format as Figure 4. Figure 11 illustrates the generation of a movie with 128 frames or 64 voxels from a structure learned from compressing a video of 64 frames (a cycle through two flaps of the wings). The RGN has compressed each cycle into eight events, repeated four times during generation. The upper right panel shows the discovered transitions among these events. In this instance, we have a simple *orbit* or closed path, where the last event transitions to the first.

The upper left panel depicts the posterior distribution over states at the highest level in image format; here, showing four cycles. These latent states then provide empirical priors over the initial states of the four image quadrants at the subordinate level (via **D**). The accompanying predictive posterior over paths (via **E**) shows that each of the four paths was constant over time; thereby generating predictive posteriors over the requisite states at the first level (i.e., singular variates) and, ultimately, the posterior predictions in pixel space. The first and last generated images are shown in the lower row.

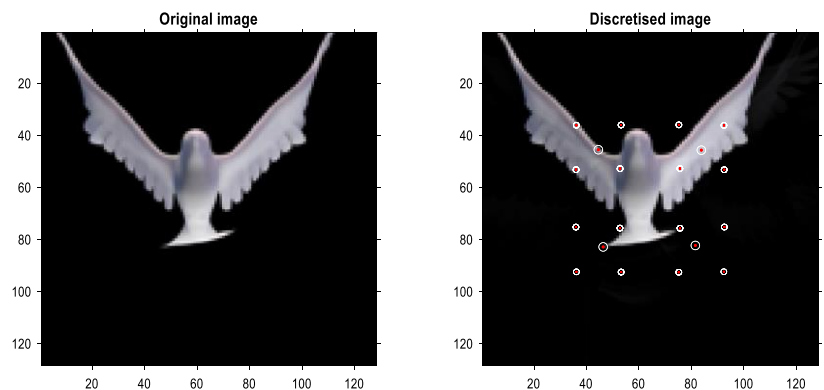

**Figure 10: A Dove in flight.** This figure shows a frame from a movie of a (digital) Dove flapping her wings. The left panel is a TrueColor (128×128 pixel, RGB) image used for structure learning, while the right panel shows the corresponding posterior prediction following discretisation. This example used a tessellation of the pixels into 32×32 voxels, with a temporal resampling of $R = 2$: i.e., successive pairs of (32×32 pixel) image patches where grouped together for singular value decomposition. Singular variates took 9 discrete values (centred on zero), for a maximum of 32 singular vectors. The locations of the image patches are shown with small red dots (encircled in white). The larger dots correspond to the centroids of blocks, following the first block transformation at the second level of the ensuing RGM.

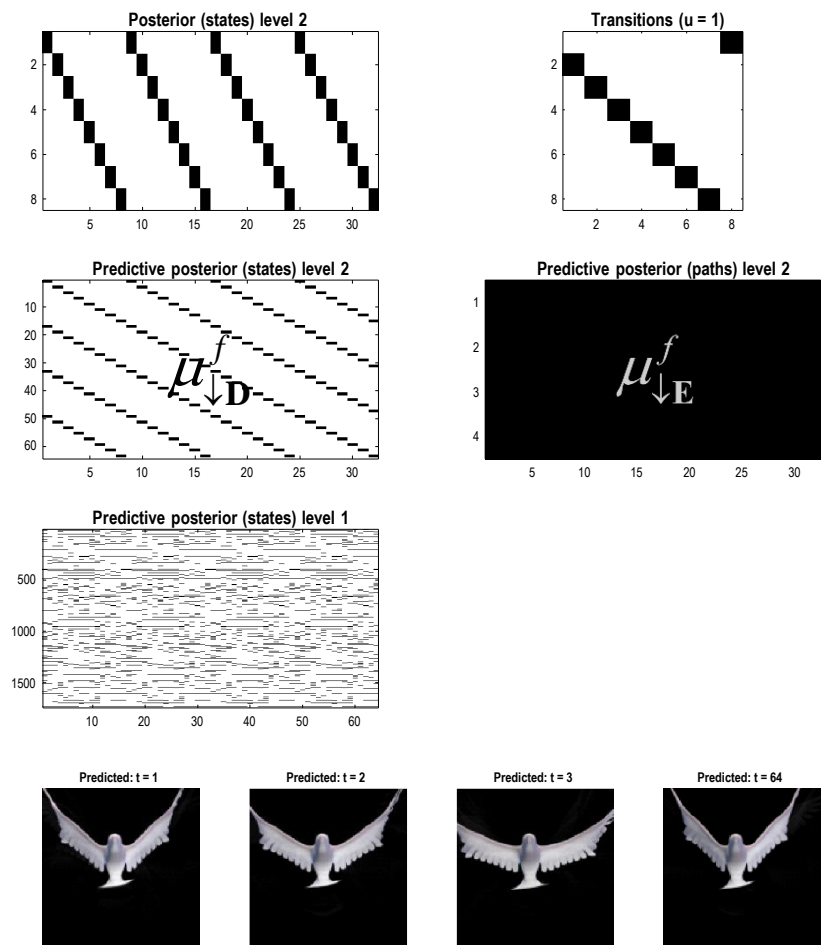

**Figure 11: Generating movies**. Following structure learning based upon two cycles of wing flapping (i.e., 64 frames or 32 time-colour-pixel voxels), an RGM was used to generate posterior predictions over 128 video frames; namely, four flaps. The structure learned under this RGM compressed each cycle into eight events. The format of this figure will be used in subsequent examples: the upper right panel shows the discovered transitions among (high level) events. In this instance, we have an orbit, where the last state transitions to the first. The upper left panel depicts the posterior distribution over states at the highest level in image format; here, showing four cycles. These latent states then provide empirical priors over 64 initial states of the four image quadrants at the subordinate level, depicted in the predictive posterior panel below. The accompanying predictive posterior over paths at this level (on the right) shows that each of the four paths was constant over time; thereby generating predictive posteriors over the requisite states at the first level (i.e., singular variates) and, ultimately, the posterior predictions in pixel space. The first and last generated images are shown in the lower row.

Figure 11 illustrates the ability of the RGM to generate video content. Conversely, Figure 12 illustrates the inference or recognition of images presented as (partial) stimuli, using the same format. The upper panels show the predictive posteriors over states (on the left) and paths (on the right), respectively. The images below correspond to the predictions in pixel space at the first time points and at the last time point. The corresponding stimuli are shown in the lower row of images. This illustration of model inversion speaks to some key biomimetic aspects.

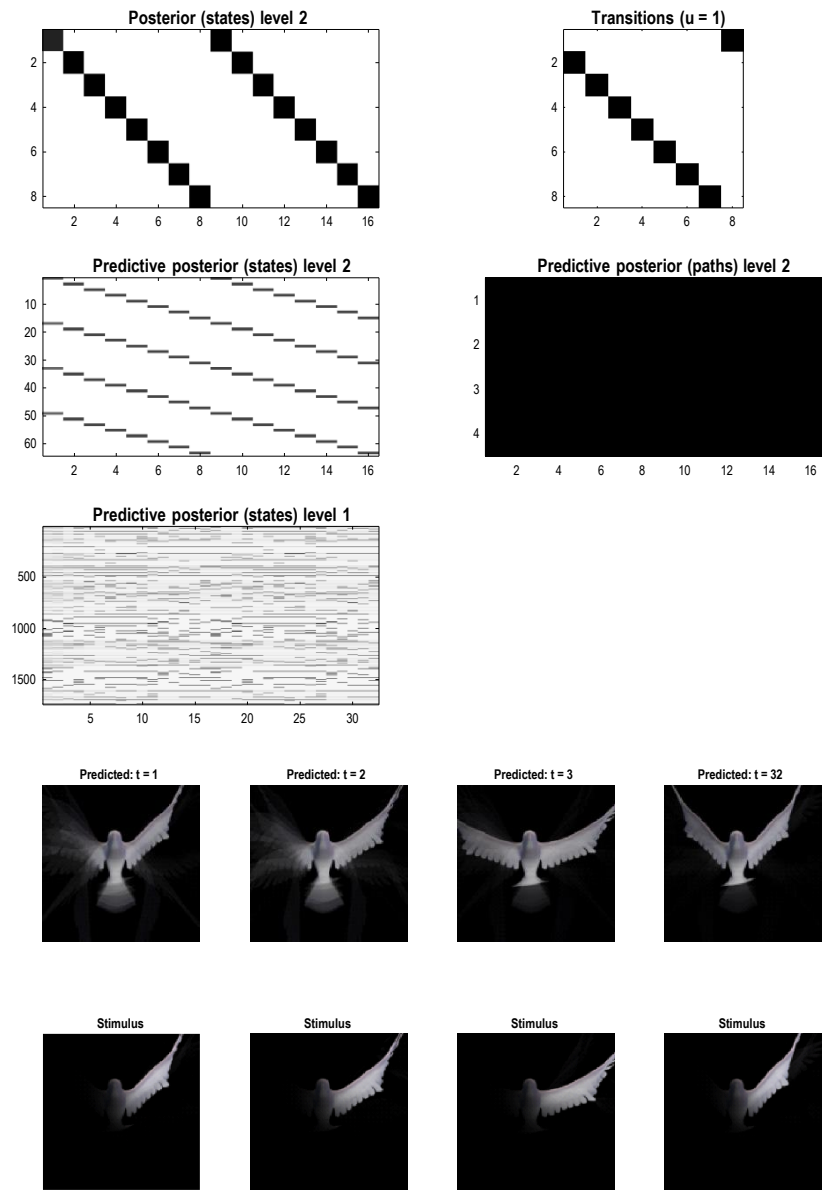

**Figure 12: Image completion**. This figure reproduces the previous figure, but presenting the model with a partial stimulus in the upper right quadrant. The likelihood mappings were equipped with small concentration parameters (of 1/32) to model any uncertainty around the events installed during structure learning. The lower rows show the posterior predictions (upper row) and stimulus (lower row) for the first and last timeframes. The key thing to take from this figure is that by the sixth video frame or third voxel (t = 3), the posterior predictive density has correctly filled in the missing quadrants—and continues to predict the stimuli veridically, by treating the missing data as imprecise or uninformative.

First, by generalising scale transformations to both space and time, we induce nontrivial posteriors over paths or dynamics. The separation into predictive posteriors over states and paths has a clear homology with the segregation of processing in the visual cortical hierarchy in the brain (Ungerleider and Mishkin, 1982). This is often cast in terms of a distinction between dorsal and ventral streams, in which the dorsal stream is concerned with 'where' things are in the visual scene and how they move—or can be moved (Goodale et al., 2004). In contrast, the ventral stream is responsible for encoding 'what' is causing visual impressions. Furthermore, higher or deeper levels of the visual hierarchy show slower stimulus-bound responses than lower levels: e.g., (Hasson et al., 2008). This is an emergent property of the (perceptual) inference demonstrated in the above example, in terms of the successive slowing of belief updating at higher levels.

In Figure 12, the stimuli presented to the RGM were restricted to the upper left quadrant. In neurobiology, this would be like presenting a moving bar in a restricted portion of the visual field (Livingstone and Hubel, 1988). Despite this partial stimulus, the top-down predictions quickly evince a form of pattern completion, effectively seeing what was not actually presented. Indeed, by the sixth frame (i.e., third voxel), the posterior predictions have filled in the missing content. This predictive capacity is reminiscent of functional magnetic resonance imaging studies of predictive processing in human subjects, in which one can record predictive activity in visual cortex that is induced by just providing one quadrant of the visual stimulus (Muckli et al., 2015).

## Paths, orbits and attractors

In the preceding example, the sequence of images constitutes a closed path; namely, a simple orbit. This can be regarded as a quantised representation of a periodic attractor. Here, we use the same procedures to compress and generate stochastic chaos, using images generated by a Lorenz system (Friston et al., 2021a; Lorenz, 1963; Ma et al., 2014; Poland, 1993). The aim of this example is to show how active learning—after structure learning—furnishes a generative model of chaotic orbits. In this instance, the dynamics or transitions are learned to accommodate switches among paths that acquire a probabilistic aspect, due to random fluctuations and exponential divergence of trajectories.

Figure 14 shows the timeseries used to generate training images. The upper right panel shows a sequence of hidden states over 1024 time bins, generated by solving stochastic differential equations based upon a Lorenz attractor. The upper left panel shows the random fluctuations (i.e., state noise)—used in generating these states—in terms of an arbitrary mixture of hidden states and stochastic fluctuations on their motion (i.e., prediction and error). The ensuing hidden states were used to generate an image sequence, in which the motion of a white circle traced out the trajectory of the first two hidden states (indicated with golden dotted lines). The first half of the resulting sequence was then used for structure learning. A training image and its reconstitution following discretisation are shown in the lower panels of Figure 14. The encircled red dots show the location of the groups of pixels at successive scales.

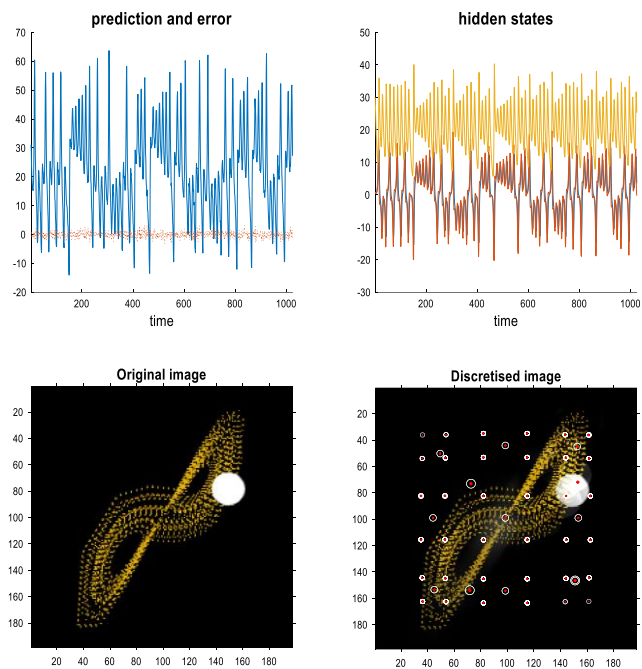

**Figure 13: Stochastic chaos**. This figure summarises the quantisation of images generated from stochastic differential equations based upon the Lorenz system. The upper panels report the solution in terms of the three hidden states of a Lorenz system (right upper panel) and the contribution of random fluctuations, innovations or state noise (left upper panel). This contribution is characterised in terms of an arbitrary linear mixture of the hidden states and the prediction errors induced by random fluctuations (red line). The hidden states were used to generate an image, in which the position of a white ball was specified by the first two hidden states. The ensuing trajectory was used to populate the image with (golden) dots. One can envisage the ensuing sequence of video frames as depicting a white particle flowing in a medium whose convection is described using the Lorenz equations of motion. [Strictly speaking, the equations pertain to the eigenmodes of convection]. The lower left panel shows an exemplar video frame in a TrueColor (198×198 pixel) image. The lower right panel shows the reconstructed image generated from its quantised representation. Following the format of Figures 4 and 10, the encircled red dots show the centroids of subsequent groups. In this example, the image was tessellated into (32×32) pixel groups with singular variates taking five discrete values for a maximum of 16 singular vectors. As previously, the temporal resampling considered successive pairs. The resulting three level RGM is illustrated in the subsequent figure.

Structure learning was based upon the first 512 images. Following this, the model was exposed to the subsequent 512 images, to enable learning of the transition dynamics via the accumulation of appropriate Dirichlet parameters. The resulting transitions are shown in the top left of Figure 15. The resulting model has represented this dynamical system with 64 events (i.e., states at the highest level) with switching among certain events that recapitulates the stochastic switching of trajectories on the underlying Lorenz attractor. Figure 15 shows the predictive posteriors over states at successive levels and predicted images in response to a stimulus. This stimulus was a 128 image sequence from the training set. Following this initial 'prompt', the stimulus was rendered imprecise for a further 128 images and then removed completely for the remainder of the simulation period.

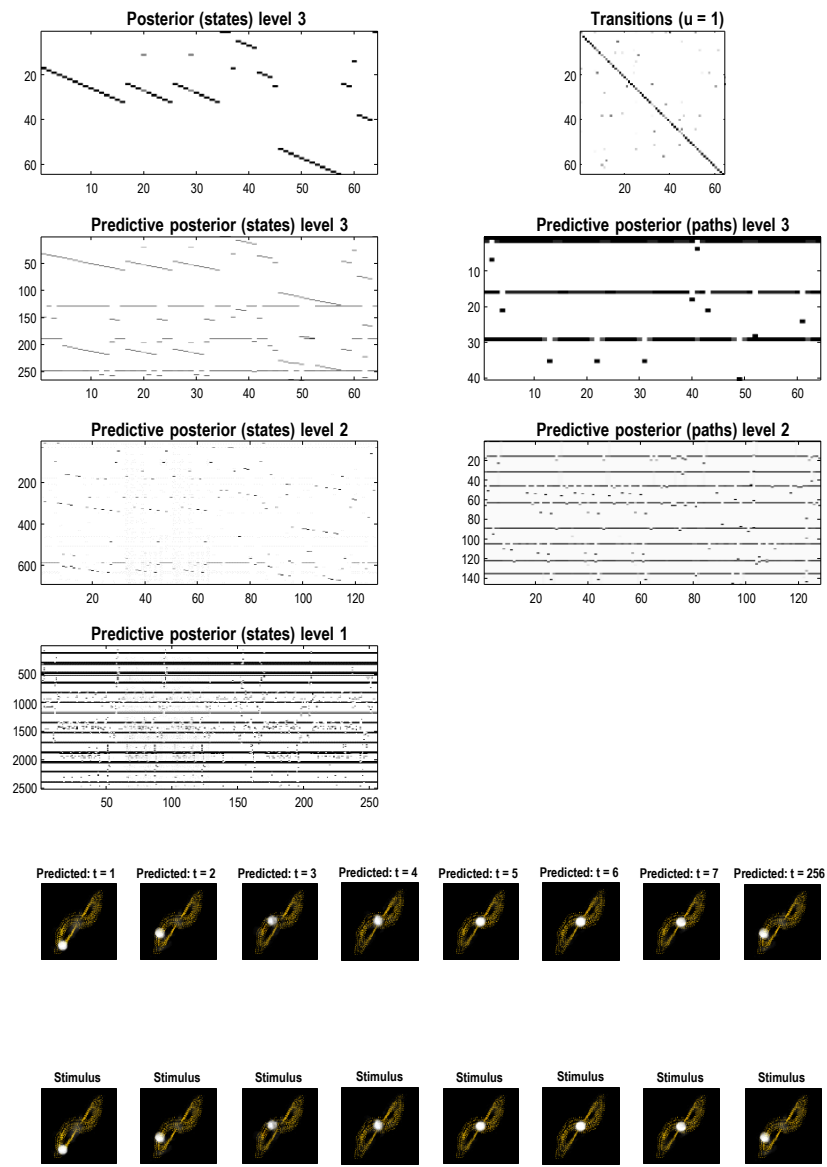

**Figure 14: Quantised stochastic chaos**. This figure uses the same format as Figure 12 to illustrate the learned transitions following fast structure learning, and subsequent active learning, based upon the first and second half of the image sequence depicted in Figure 13. In this example, the dynamics are summarised in terms of 64 events that pursue stochastic orbits, under the discovered probability transition matrix shown on the upper left. The lower panels show the posterior predictions in pixel space and the accompanying stimuli presented for the first quarter of the simulated recognition and generation illustrated in Figure 15.

Figure 15 shows the stimuli and posterior predictions induced with this stimulation protocol. Maximum intensity projections—over the horizontal dimension—were concatenated to render the fluctuations of the first hidden (Lorenz) state easily visible (compare Figure 15 with Figure 14). This format shows that the first 128 images have been recognised veridically; namely, the paths through renormalised latent state-spaces have been correctly identified. When the stimulus is rendered imprecise (dark region in the lower panel), the posterior predictions continue to produce plausible chaotic dynamics, until the stimulus is removed altogether, at 256 time bins. At this point, the RGM generates its own outcomes, based upon the learned generative model, which correctly infers the latent causes of this self-generated content. In other words, the second sequence of stimuli in the lower panel of Figure 16 are generated by the model's *discrete* representation of chaotic events, as opposed to the first period, in which they were generated by solving *continuous* stochastic differential equations.

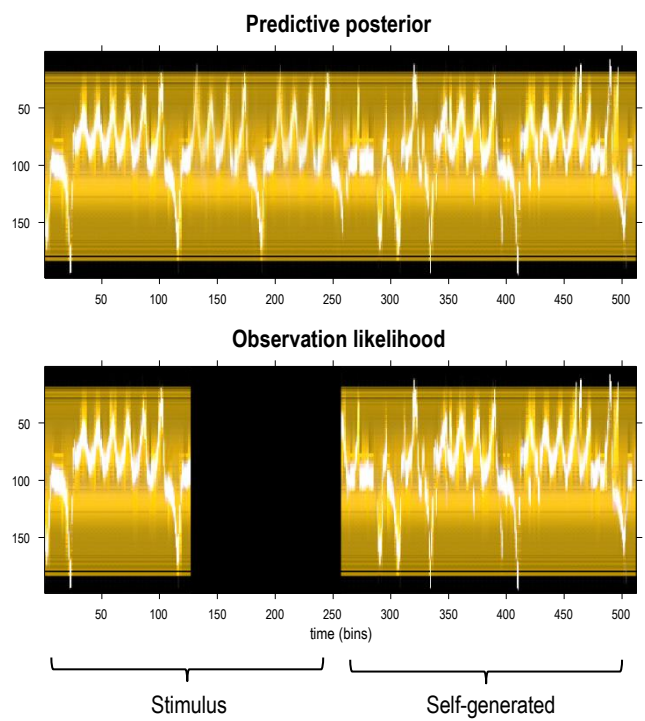

**Figure 15: Generating stochastic chaos.** This figure illustrates the sequence of images predicted (and presented) based upon the posterior predictive distributions of the previous figure. Here, maximum intensity projections of each frame have been concatenated to show the video as a single image (i.e., as if each image were viewed from the side). The upper panel shows the predictive posterior in pixel space, while the lower panel reports the stimulus presented to the model. Crucially, the first quarter of the stimulation used images generated from the Lorenz system, while the second half of the stimulus was self-generated; namely, sampled from the learned generative model. The intervening dark regime (second quarter) denotes a period in which the input was rendered imprecise (i.e., presented with poor illumination or with the eyes shut). Despite this imprecise input, the posterior predictions continue to generate plausible and chaotic dynamics, until they become entrained by self-generated observations. Here, the simulations lasted for 512 frames (i.e., time bins).

These numerical studies speak to two interesting points. First, the RGM can be regarded as the discrete homologue of switching linear and nonlinear dynamical systems; e.g., (Linderman et al., 2016; Olier et al., 2013; Rabinovich et al., 2010; Tani and Nolfi, 1999). However, in this quantised setting, there is no distinction between linear and nonlinear. This follows because all metric state-spaces have been tessellated and coarse-grained. This means that the only metric space is the statistical manifold (simplex) associated with probability distributions over discrete events. In principle, these discrete models can handle any degree of nonlinearity in continuous state-space models.

The second issue is a perspective on embedding (Deyle and Sugihara, 2011; Klimovskaia et al., 2020; Ørstavik and Stark, 1998). One might ask how the RGM can recover chaotic dynamics from a sequence of static images. The answer lies in Takens embedding theorem (Deyle and Sugihara, 2011; Takens, 1980), which means that any (chaotic) attractor can be reconstructed from a time-delay embedding, which is implicit in the temporal RG operators used in renormalisation. Effectively, we started with an infinite dimensional system (due to the inclusion of random fluctuations) and projected a realisation into high-dimensional pixel space. Because the resampling discretised successive pairs of images, we are effectively summarising the state and the velocity at each (group of) pixels at the first level of the RGM, which subsequently embeds pairs of pairs, during renormalisation. At the highest level, we are left with a probabilistic representation of chaotic dynamics on a 64-dimensional simplex, which retains the essential structure of the random dynamical attractor. Next, we will consider a more natural random dynamical attractor, afforded by the motion of natural kinds.

## A natural extension

Figure 16 illustrates the spatial blocking of a more natural video—a short sequence (from pixabay.com/video/search/birds) in which a Robin alights on a branch, feeds and preens itself for a few seconds and then flies away. In this example, each block comprises 16×16 pixels, tessellating (128) video frames, cropped to 128×128 pixels. During fast structure learning, this video was compressed—via three renormalisations—under a three level RGM. Figure 17 illustrates the posterior predictions of each level, on exposure to a segment of the original movie. Note, in comparison with the previous example, the dynamics are more itinerant. This reflects the fact that this (naturalistic) video has more frequent transitions, as the Robin repeated stereotyped behavioural repertoires.

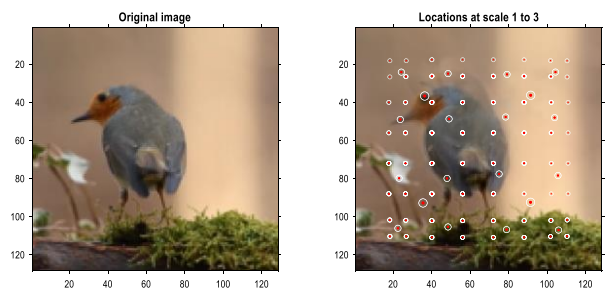

**Figure 16: natural kinds**. This figure illustrates a single video frame from a short movie of a bird feeding and preening. This movie sequence comprised 128 frames of (128×128) TrueColor images. Following the format of previous figures, the left panel shows an original image and the right panel the corresponding image generated from discrete singular variates. In this example, the singular variants took 17 discrete values for a maximum of 32 singular vectors. As previously, the temporal scaling was $R = 2$; i.e., pairs of video frames were grouped together to constitute time-colour-pixel voxels. The locations of successively grouped voxels are shown with encircled red dots engendering the three level RGM reported in Figure 17.

The lower panels of Figure 17 show that the posterior predictions in pixel space were quickly entrained by the stimulus or observations, to the extent that the bird's movements were correctly predicted, despite withholding frames during movement initiation. This ability to 'fill in the gaps' reflects the 'filling in' of quadrants in the previous (Dove) example. However here, the filling in is over time, as opposed to pixel space. This ability to generate sequences is reminiscent of generative AI, in which we can regard the initial frames as a 'prompt' that is sufficient to generate a succession of future frames. However, in this setting, this predictive capacity does not rest upon an autoregressive model of the kind used in transformers (i.e., mapping past content to future content): rather, there is an explicit generative model of trajectories or paths—due to separation of temporal scales—which are, effectively, entrained by observations or content. In the next section, we pursue the temporality of this class of models by applying the same procedures to birdsong and music.

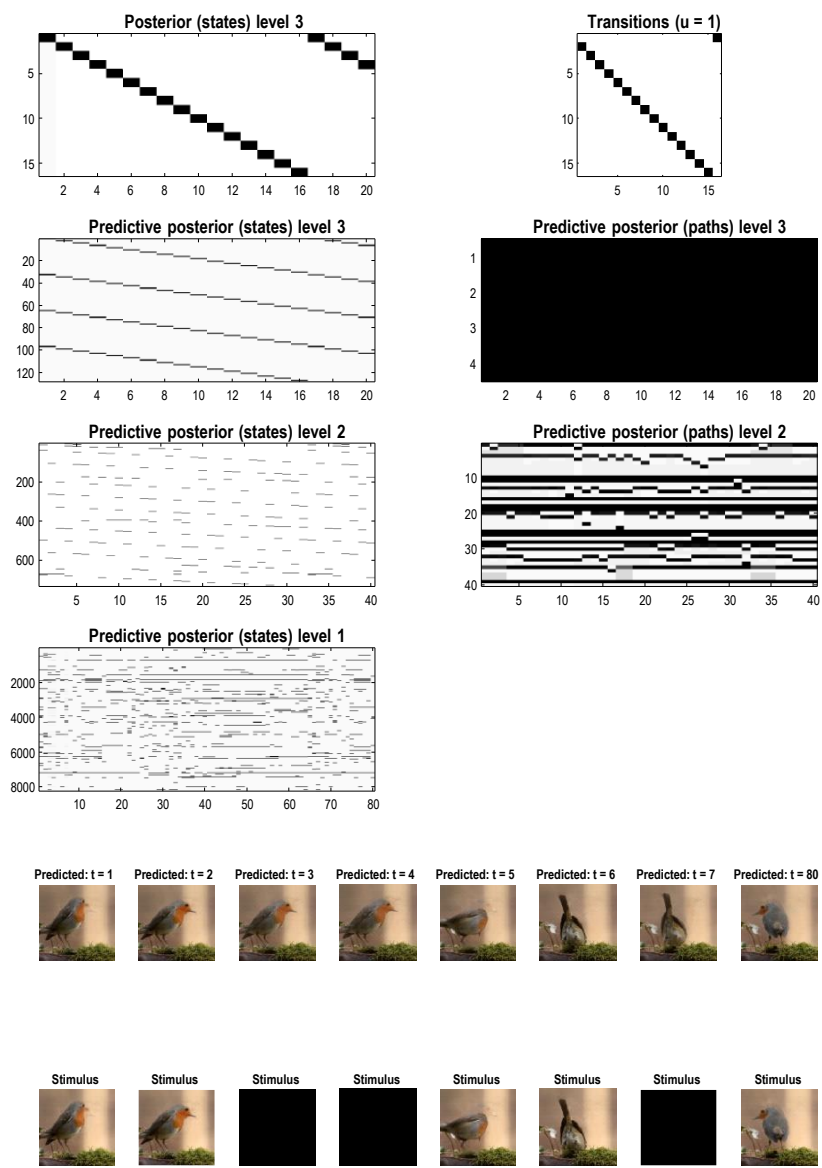

**Figure 17: Now you see it. Now you don't**. This figure uses the same format as Figure 12 to illustrate the recognition and generation of a short movie—of a Robin—comprising 160 video frames (where the last 32 frames were a repeat of the first 32). The reason there are only 80 time-steps on the *x*-axis for the level 1 predictive posteriors is that these predictions only pertain to the first of a pair of video frames. The implicit loop has been summarised as a simple orbit through 16 events. In this example, a stimulus was presented for the first four frames and then removed for the subsequent four frames. Despite the absence of precise stimuli, the posterior predictions veridically track the motion of the bird during the missing stimulus.

## SOUND COMPRESSION AND GENERATIVE AI

In this section, we focus on an exemplar application to auditory streams or sound files. In this setting, pixels (i.e., picture elements) are replaced by voxels (volume elements) over frequency and time. These constitute a time-frequency representation of timeseries; e.g., a continuous wavelet transform (CWT). Here, timeseries are mapped to time-frequency space using (Morlet) wavelet transforms, while an inverse transform converts CWT representations to a linear sound file for playing. Renormalising generative models for sound are simpler than for video content, because there is only one metric dimension (i.e., frequency) that accompanies time. Using a (spin) block transformation to coarse-grain over frequencies reflects the fact—or assumption—that neighbouring frequencies fluctuate in a correlated fashion.

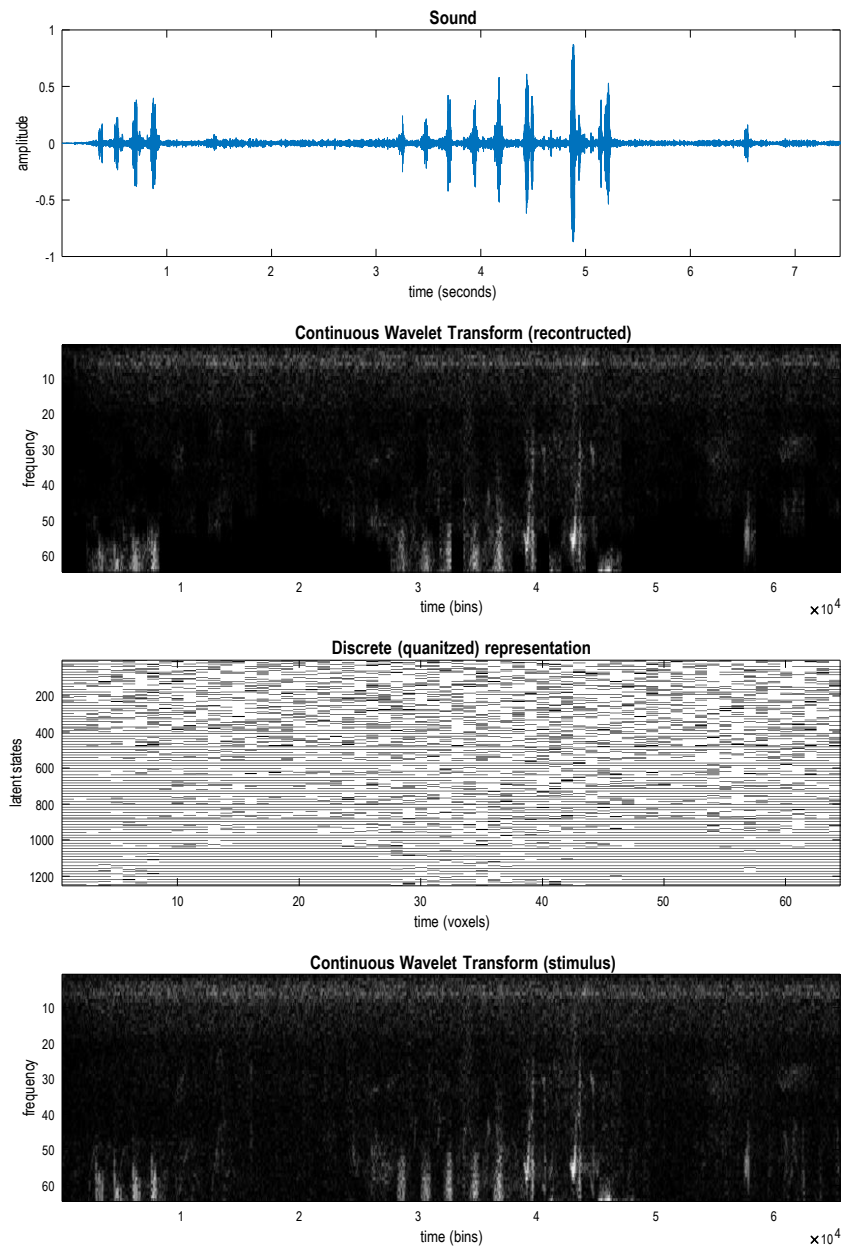

**Figure 18: Sound images**. This figure shows the training data for structure learning. The lower panel shows the continuous wavelet transform (CWT) of a recording of two (Crossbill) bird calls, where each call comprises a crescendo of short chirps. The CWT is shown as an image of time frequency responses; i.e., the spectral power from 1 to 64 frequency bins, as it evolves over time. Strictly speaking, this is not a continuous wavelet transform because the Gaussian envelope of the Morlet wavelets was fixed at 32 ms. As such, this is effectively a short-term Fourier transform between 40 and 4000 Hz (apt for the frequency range of human hearing). The second panel

shows the equivalent representation generated from the quantised representation in the third panel. The discrete representation is shown as an image of the probability distributions over singular variates associated with the singular vectors (i.e., time-frequency basis sets) used for discretisation. The upper panel shows a sound file, generated from the (reconstructed) CWT.

## A worked example

We will first look at a simple example using sound recordings[13] of a bird (a Crossbill) and then look at a more deeply structured, semi-Markovian process; namely, jazz music. Figure 18 shows the content modelled using the same procedures above, after quantising into 64 frequency bins, with a maximum of 16 singular vectors for each group of (1024×4) time-frequency voxels. The corresponding timeseries is shown in the upper panel, sampled from the CWT in the lower panel. In this example, lasting for about eight seconds, the bird makes two calls, each comprising a short sequence of chirps. The second call is longer, with a crescendo of increasingly broadband chirps.

---

[13] Copyright free files were obtained from the following website: https://sound-effects.bbcrewind.co.uk/

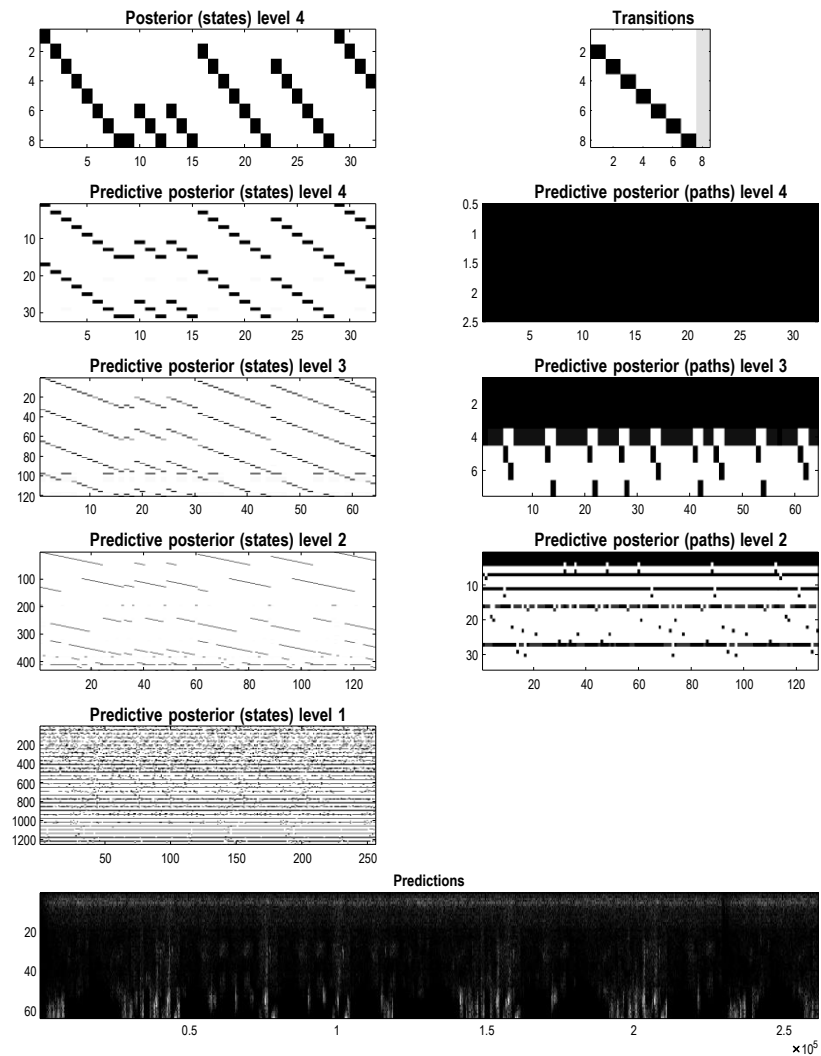

**Figure 19. Renormalising birdsong.** This figure uses the same format as Figure 17 but displays the posterior predictions as a continuous wavelet transform (i.e., a time frequency representation of the spectral power at each point in time). In this example, time-frequency voxels covered four frequency bins and 1024 time bins; corresponding to about 100 ms (at a sampling rate of 8820 Hz). Quantisation of the ensuing time-frequency voxels used singular variates with five discrete values, for a maximum of 16 singular vectors. This example reports the generation of birdsong after compression to a sequence of eight events. Here, the final event could be followed by any preceding event with equal probability. This follows because there was no recurrence of events in the training data used for structure learning. As a consequence, during generation, events cascade to the eighth event and then transition to a preceding event stochastically. The upper left panel shows the resulting succession of events that produce the posterior predictive sequence of bird calls in the lower panel. When played, the resulting sound file is indistinguishable from a bird emitting a variety of stereotyped calls, in a quasi-random sequence.

After fast structure learning, this recording engendered an RGM with four levels, with highest level states encoding an event of about one second, where episodes included intervening periods of a low-frequency background rumble (visible as a band of low-frequency, continuous power in the CWT). This RGM can now be used to generate arbitrarily long sequences of calls. An example is shown in Figure 19, over about 32 seconds (or 256 time-frequency voxels). The upper right panel shows that the timeseries has been compressed into a succession of eight episodes that follow each other systematically until the last; after which the model has uniform beliefs about the next event. This is reflected in generated birdsong, in which the RGM completed a sequence of eight events, repeated the last event and then selected the sixth event, and so on. This resulted in alternating calls, as seen in the sonogram in the lower panel.

We could have continued generating calls indefinitely, where each successive episode would be longer or shorter, depending upon the point of 're-entry'. There are clearly many other variations on this theme; for example, we could have implemented a transition from every event to the first event, so that sequences terminate at varying times—or terminate on particular events (e.g., as in language). The question now is: would a different sequence be predicted in the presence of auditory input?

To illustrate the dual role of generation and recognition, under these models, we repeated the simulation above but presented auditory input, starting with the second call. See Figure 20. In this instance, the posterior over the initial episode now correctly begins during the onset of the second call and faithfully predicts the auditory input for the duration of its presentation (here, a couple of seconds). After this initial period of stimulation, the auditory input was removed (by delivering outcomes with an imprecise, uniform probability distribution). The RGM effectively treats the stimulus as a 'prompt' and pursues its itinerant cycle of song generation.

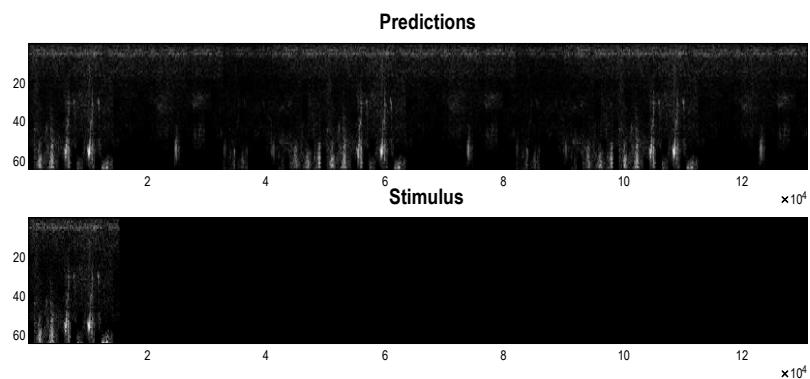

**Figure 20: Song recognition**. These time-frequency representations reproduce the lower panel of Figure 19 but in the context of an initial stimulus or 'prompt', corresponding to the second call in the training data. These posterior predictions show that the model immediately identified the 'call' and its phase. And then continued to generate predictions, in accord with its model of successive auditory events. In this example, the stimulus was rendered imprecise (i.e., inaudible) after 16 of the 128 time-frequency voxels generated.

The preceding illustration shows that content is assimilated and predicted (i.e., generated) online, without the need for caching. This follows because the Bayesian belief updating has, by design, a scheduling that eludes backwards message passing in time (a.k.a., Bayesian smoothing). This kind of reactive message passing can be conceived of as follows: the highest level sends a prior message to the level below. The level below reacts to this top-down message by performing two updates before returning a likelihood message requested by the higher level. The higher level forms a posterior belief on the basis of the likelihood message and updates its posterior belief about the current *and next state*, which it then provides as an empirical prior to the level below, and so on. As noted in previous sections, any level can only update its beliefs after waiting for its subordinate levels to accumulate two messages. In short, the posterior beliefs at any given level are retrospective beliefs, after accumulating evidence from lower levels over the timescale in question. The ensuing predictive posteriors are then used as priors to provide contextual constraints on the initial states and paths of lower levels—and likelihoods for the states at the level above. In the MATLAB implementation of these demonstration routines, this reactive message passing emerges from the nested composition of function calls. It is interesting to consider how the requisite scheduling would be implemented with reactive message passing (Bagaev and de Vries, 2021); c.f., the actor model (Hewitt et al., 1973) (Keith Duggar; personal communication).

## From songs to music

One might ask if RGMs could be applied to language, perhaps in the spirit of hierarchical Dirichlet process models: e.g., (MacKay and Peto, 2008; Teh et al., 2006). We will not address this here but provide an application to music to illustrate the generalised synchrony that accompanies communicative exchange (Friston and Frith, 2015). Figure 21 shows the sound file used for fast structure learning. In this example, a short (two minutes) sound file of a jazz pianist was discretised to 64 time-frequency voxels (between 40 and 4000 Hz in 32 bins). The ensuing generation of piano music is shown in Figure 22, where the RGM has compressed the sound file into 16 events, each corresponding to a bar of music. In this example, the model was exposed to a succession of music segments using active learning— Equation (7)—after fast structure learning (Algorithm 1). In virtue of the timeseries presented, it learnt that the last event (i.e., musical bar) was followed by the first or eighth bar of music. The RGM now generates extended piano play, selecting successive eight-bar sequences, in accord with the learnt statistics of variation at this timescale.

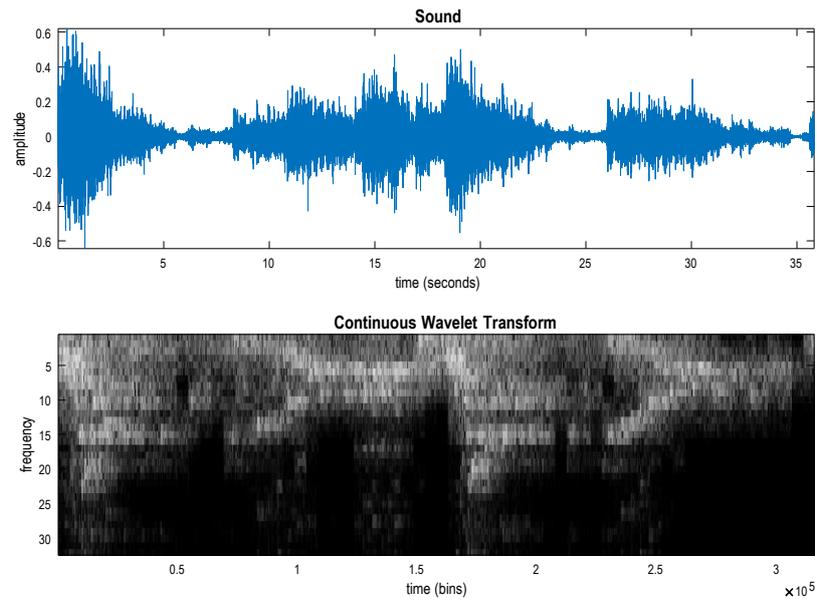

**Figure 21: jazz music**. This figure follows the same format as Figure 18; however, here, the recording is of 36 seconds of jazz piano, comprising about 16 bars. The continuous wavelet transform was quantised into 32 frequency bins between 40 and 4000 Hz (with a fixed Gaussian envelope of 8 ms). The (Nyquist) sample rate was twice the highest frequency considered. The time-frequency representation was quantised using time-frequency voxels of four neighbouring frequencies and a time bins covering about 500 ms, corresponding to 1/4 of a musical bar. Following renormalisation, musical events at the highest (third) level has a duration of 2.24 seconds; i.e., a bar of music.

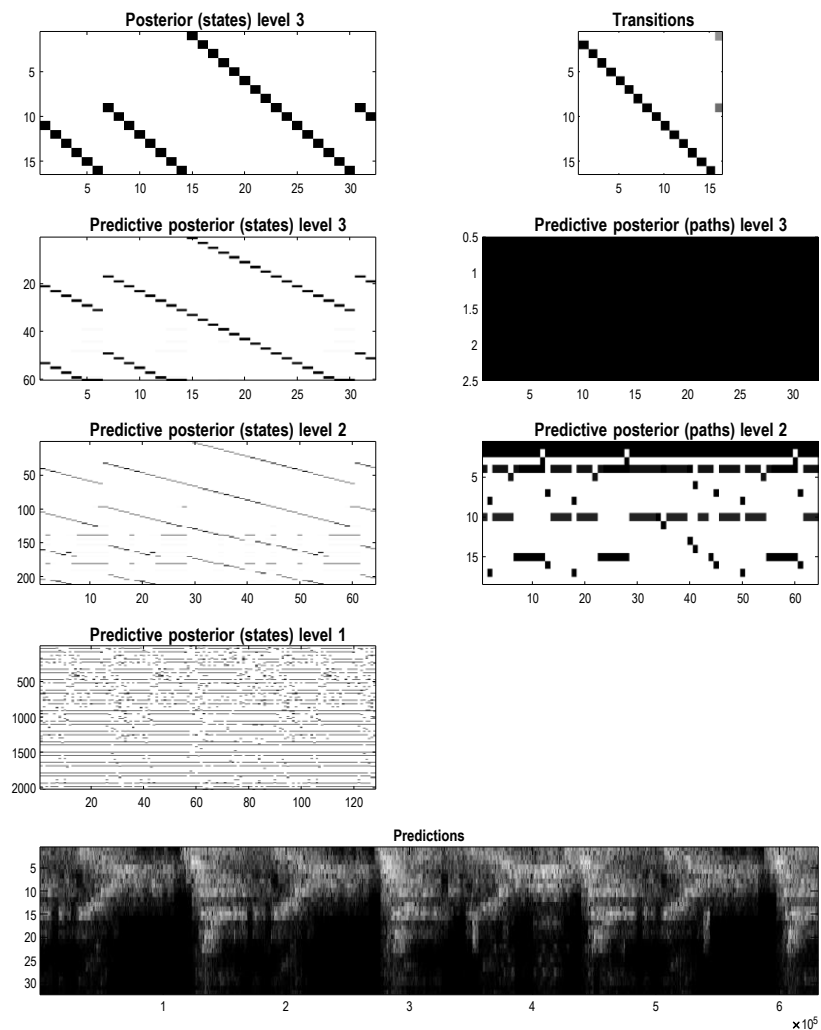

**Figure 22: Generating music**. This figure follows the same format as Figure 19. In this example, each event corresponds to a bar of music and the simulation reports generation of 32 bars, under the learned transitions shown on the upper right. Here, the model has learned to generate eight bars of music until the final bar; after which re-enters at the first or ninth event to pursue its path through event space. In other words, the model generates stochastically alternating eight-bar musical sequences. This particular generative behaviour is a simple consequence of what it has heard during structure learning and subsequent active learning.

Figure 23 shows the same simulation during the presentation of the original sound file. The key thing to note is that the predictions are almost instantaneously synchronised with the heard music. If we read the predictions as the sounds that would be generated by an agent—and the stimuli as the music generated by an accompanist—can regard this simulation as a discrete analogue of the generalised synchronisation that emerges in dyadic interactions, under a shared generative model; i.e., singing from the same hymn sheet (Friston and Frith, 2015).

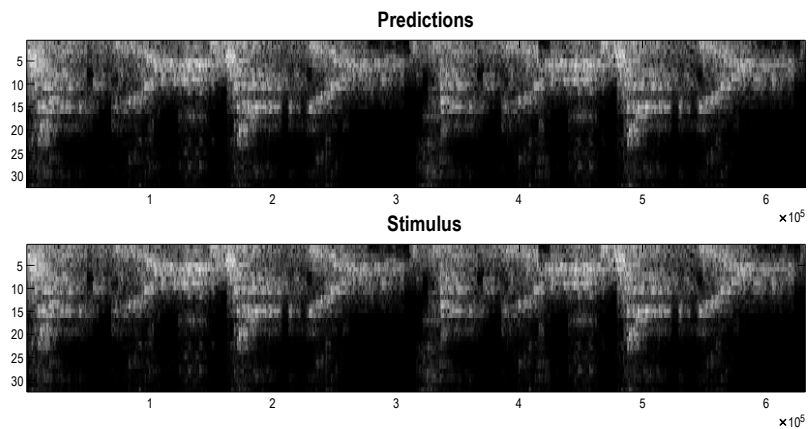

**Figure 23: Musical accompaniment**. This figure follows the same format as Figure 20 and illustrates the synchronous entrainment of posterior predictions by stimuli. This entrainment can be read as a generalised synchrony (a.k.a., synchronisation of chaos) under a shared musical narrative or generative model.

# FROM PIXELS TO PLANNING

In this final section, we turn to the deployment of RGMs for planning as inference (Attias, 2003; Botvinick and Toussaint, 2012; Da Costa et al., 2020) and, implicitly, their use as synthetic agents and decision-makers under uncertainty. This application is relatively straightforward under active inference, due to its roots in the free energy principle and implicit biomimetic commitments. From the perspective of the free energy principle, agents are read as certain kinds of self-organising systems that possess characteristic states, which characterise the kind of agent or system they are. The very existence of this attracting set or manifold (a.k.a., pullback attractor) means that one can describe or simulate self-organisation as a variational principle of least action (Friston et al., 2023a). Application of this principle underwrite the variational belief updating of Figure 2.

The notion of an attracting set is useful here because it foregrounds the role of constraints; in the sense there are certain states—outside the attracting set—that an agent is unlikely to be found in. This means the attracting set can be described in terms of prior preferences, whose logarithm can be interpreted as value (for the agent in question). The negative logarithm corresponds to self-information; providing an information theoretic account of self-organisation in terms of value-pointing dynamics. From the current perspective, the self-information is the quantity bounded by variational free energy.

From the perspective of reinforcement learning, one can regard sparse rewards as specifying unstable fixed points on the pullback attractor. This means that if we wanted to learn the structure of a generative model of a particular (expert) agent, we simply need to compress the paths of least action that link rewarding states. Equipped with such a model, action—under active inference—simply realises the predictions under such a model; thereby realising expert play or Bayes optimal decision-making under uncertainty: see Equation (3).

From a biomimetic perspective, this replaces the notion of motor commands with motor predictions that are fulfilled by peripheral motor reflexes (Adams et al., 2013; Friston, 2011). This view of motor (and autonomic) control inherits from the partitioning of states under the free energy principle. In brief, internal states—e.g., of a neuronal network—are separated from external states—e.g., of the body or extrapersonal space—by control and sensory states that, together, constitute blanket states. This leads to a form of active inference that can be regarded as control as inference (c.f., model predictive control) in which internal states generate predictions or setpoints that are realised reflexively by active states: c.f., (Kappen et al., 2012). Crucially, both internal and control states can be cast as minimising variational free energy. For internal states this manifests as inference and learning. For control states this reduces to minimising proprioceptive prediction errors[14] in motor control (Friston et al., 2011). Or telemetry prediction errors in drones or other artefacts (Lanillos et al., 2021).

This kind of active (control as) inference can be usefully contrasted with the optimisation of state-action policies in reinforcement learning. Figure 24 illustrates this in terms of the differences in computational architectures, which focus on planning as inference (i.e., sequential policy optimisation) and learning to plan (i.e., learning state-action policies), respectively.

---

[14] Where prediction errors can be read as variational free energy gradients.

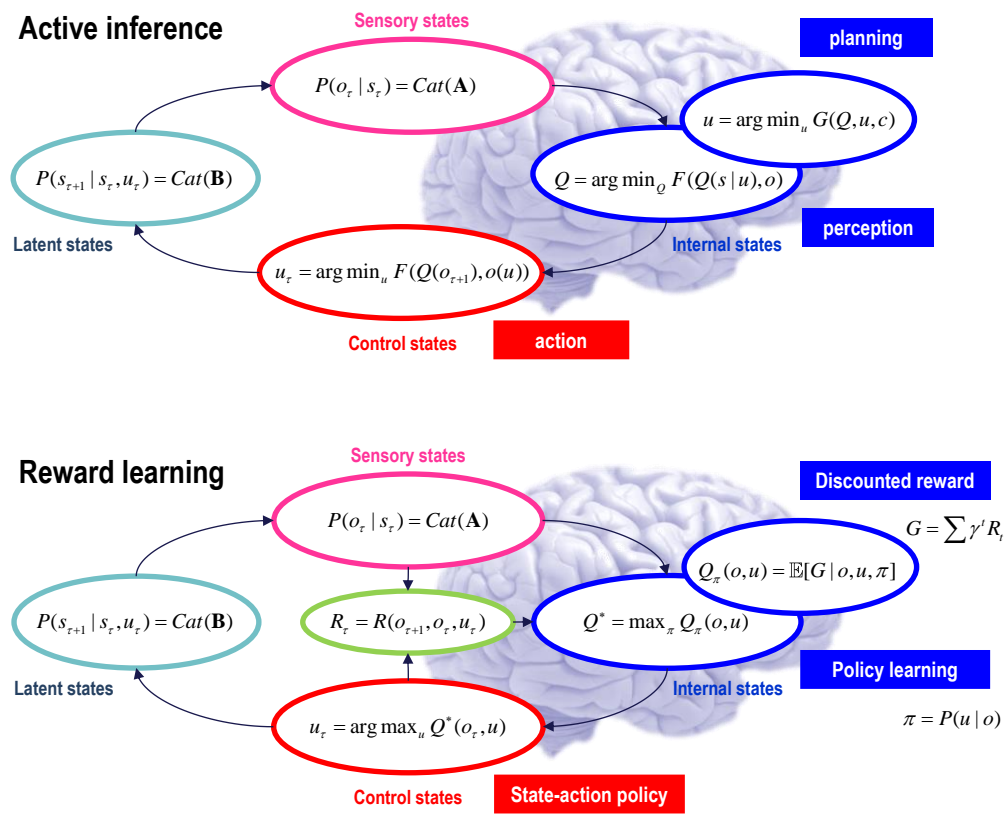

**Figure 24: Active inference and reinforcement learning**. This figure provides two schematics to highlight the difference between active inference and reinforcement learning (i.e., reward learning) paradigms. Active inference can be read here as subsuming a variety of biomimetic schemes in control theory and the life sciences; such as control as inference (Kappen et al., 2012), model predictive control (Schwenzer et al., 2021) and—in neurobiology—motor control theory (Friston, 2011; Todorov and Jordan, 2002), perceptual control theory (Mansell, 2011), the equilibrium point hypothesis (Feldman, 2009), *etc.* The basic distinction between active inference and reinforcement learning is that in active inference, action is specified by the posterior predictions in outcome modalities reporting the consequences of action. These posterior predictions inherit from policies or plans that minimise expected free energy; namely, Bayesian planning as inference (Attias, 2003; Botvinick and Toussaint, 2012; Da Costa et al., 2020). This kind of planning is Bayes optimal in a dual sense: it conforms to the principles of optimum Bayesian design (Lindley, 1956) and Bayesian decision theory (Berger, 2011) via the maximisation of expected information gain and expected value, respectively (where expected value is defined in terms of prior preferences). Mechanically, this can be expressed as belief updating under a suitable generative model (i.e., planning as inference) to provide posterior predictions that are fulfilled by action (i.e., control as inference). On this view, both belief updating (i.e., perception) and motor control (i.e., action) can be read as minimising variational free energy. This can be contrasted with reinforcement learning, in which there is an assumed reward function that has a privileged role in updating the parameters of a universal function approximator (e.g., a deep neural network) mapping from inputs (i.e., sensory states) to outputs (i.e., control states). The example of reinforcement learning here uses state-action policy learning based upon discounted reward: c.f., (Lillicrap et al., 2015; Watkins and Dayan, 1992).

## Planning as inference

Much of reinforcement learning—e.g., deep RL—is concerned with a difficult problem of learning state-action policies through learning the parameters of a neural network that maximise expected reward, where sparse rewards are only available after some suitable sequence of actions (Kaiser et al., 2019). Model free approaches (such as illustrated in Figure 24) learn a mapping from states to actions; while model-based schemes learn a generative model to finesse action selection (Theodorou et al., 2010).

Active model selection offers a different perspective on the requisite learning: instead of updating model parameters following a reward, one can simply update the model *per se*. In other words, one can learn a compressed representation of events that intervene between one reward and the next. This casts reinforcement learning as a structure learning problem, in which the structure is only updated if unique events do not entail a cost: i.e., if the training sequence ends with a preferred (rewarded) outcome. Equation (3) formalises this notion in terms of expected free energy—i.e., expected information gain and cost—where selection (c.f., Maxwell's Demon) is effectively applied to data used for structure learning, as in Equation (9).

Procedurally, this looks very much like conventional reinforcement learning; however, it is simpler and more efficient because it learns a compressed representation of, and only of, paths that link rewarded states. Another way of looking at this is as a 'smart data selection'. Effectively, this leads to the automatic selection of model structures that can only recognise and predict rewarding episodes, thereby eluding subsequent parameter learning. This rests upon the fact that one does not need to learn how to maximise reward if one has a generative model that can only predict paths that lead to reward—and thereby specify the next action. In what follows, we will illustrate this approach in sequential policy optimisation and unpack some of its corollaries.

## Games and attractors

To illustrate the use of the RGM for planning as inference, this section uses simple Atari-like games to show how a model of expert play self-assembles, given a sequence of outcomes under random actions. We illustrate the details using a simple game and then apply the same procedures to a slightly more challenging game.

The simple game in question was a game of Pong, in which the paths of a ball were coarse-grained to 12×9 blocks of 32×32 RGB pixels. 1,024 frames of random play were selected that (i) started from a previously rewarded outcome, (ii) ended in a subsequent hit and (iii) did not contain any misses. In short, we used rewards for, and only for, data selection. The training frames were selected from 21,280 frames, generated under random play. The sequence of training frames was renormalised to create an RGM. This fast structure learning took about 18 seconds on a personal computer. The resulting generative model is, effectively, a predictor of expert play because it has only compressed paths that intervene between rewarded outcomes. Figure 25 (lower panels) illustrates these high dimensional orbits by plotting the paths in the first few principal dimensions of the underlying statistical manifold. The rewarded states (hitting the ball) are encircled in red, while the blue lines show the multiplicity of paths that connect rewarded outcomes.

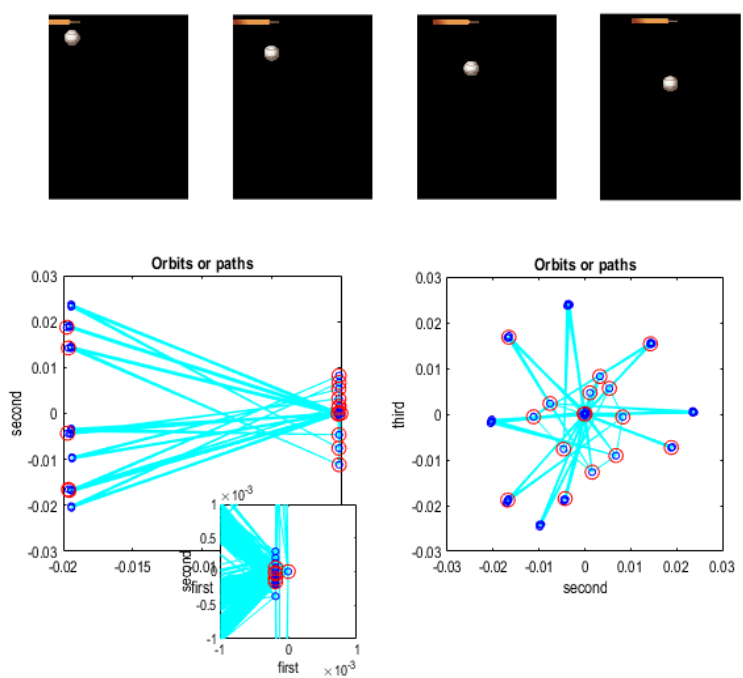

**Figure 25: Policies and attractors**. The upper panels show the first four frames generated by a game engine simulating a simple version of Pong. In this game, the paddle has to return a ball that bounces around, inside a rectangular box. These images were generated from discrete factors generating (32×32) groups of TrueColor pixels were each (12×9) factor (i.e., location) could be in five states; corresponding to three parts of the paddle, a ball or background. In the game engine or generative process, the ball simply bounced around with constant momentum, moving from one location to the next at every time step. The paddle could move in either direction by one location, or stay still. A training set of such images (in discretised space) was generated by concatenating sequences of random play that intervened between ball hits. When expressed in terms of probability distributions over quantised states, the ensuing trajectory corresponds to an orbit on a high-dimensional statistical manifold (i.e., simplex). The lower panels illustrate the itinerant nature of this orbit, in the space spanned by the first two pairs of singular vectors of the associated timeseries. The inset provides a magnification of the orbit near the origin of the projection. This inset speaks to a self-similar aspect to the transitions among unique points in this quantised (probabilistic) representation. The number of points correspond to the unique combination of image features in the training set. These points constitute an attracting set that can be learned, under an RGM. Rewarded states or configurations are circled in red; illustrating the fact that there are several paths available for getting from one sparse reward to the next (via unrewarded states).

Figure 26 shows how these orbits have been assimilated into the RGM, in terms of the transitions among events at the highest (third) hierarchical level. Where each event lasts for $4 = 2^{(3-1)}$ timesteps at the first level. The RGM compressed all such events encountered during structure learning into slightly less than the total number of events experienced: $233 < 256 = 1024/4$. The right panel shows unique transitions among events where each event is followed by one and only one subsequent event. There are three such unique transition matrices. This means that each event can be followed by up to three other events. This reflects the plurality of paths among rewarding states seen in Figure 25. In other words, there are several paths from one reward to the next that have been assimilated due to the random nature of actions subtending these paths.

Using these allowable transitions, one can identify the events that precede a rewarding event. By repeating this process recursively (and retaining proceeding events) one can work backwards in time and identify at which point there always exists a path to a rewarding event—defined as an event that contains a hit at the first level. The left panel of Figure 26 illustrates this analysis graphically, by plotting the events that include or lead to a subsequent event that is, ultimately, rewarding. It can be seen that in this example, a rewarding event is accessible in 16 events or less from most latent states at the highest level.

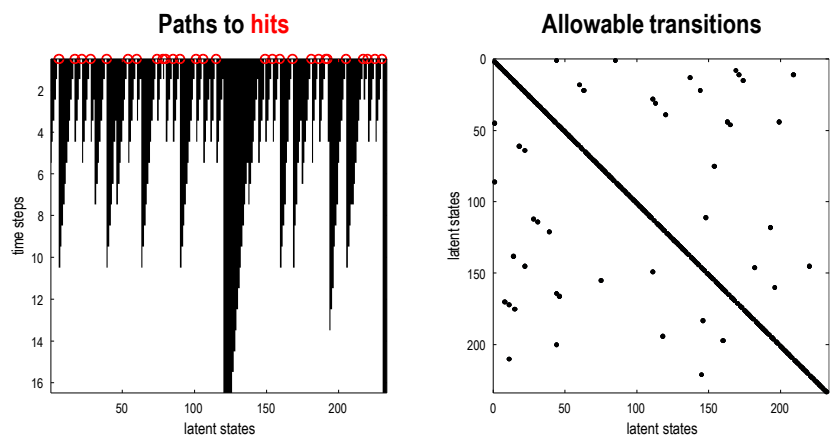

**Figure 26: Paths to success**. This figure illustrates the transitions among events following structure learning and implicit renormalisation of the training sequence. The right panel shows that the training sequence has been compressed to about 256 successive events, with occasional opportunities to switch paths. This follows from the alternative ways in which the paddle can move to reach the same (rewarded) endpoint. The left panel illustrates the paths to hits—i.e., rewarded events that include a hit (red circles)—based upon the discovered transitions on the right. The requisite paths are shown in white, while the black regions depict events that preclude a hit within the number of timesteps along the Y axis. These can be identified by simply iterating the transitions and asking whether there is an allowable transition from any given state to a rewarded state. This representation suggests that—for nearly every latent state—there is a path to a reward in 16 events (i.e., timesteps) or less. These paths are identified by inductive inference, which assigns a high cost to latent states that preclude a rewarded outcome.

One can leverage this characterisation of event-based encoding to implement *inductive inference* (Friston et al., 2023c); namely, an efficient form of active inference for discrete state-space models under precise beliefs about transitions. In brief, this involves recursive applications of the backward transitions to goal states, to identify plausible paths into the future; namely, avoiding the black states in the left panel of Figure 26. This kind of inductive planning effectively defines the paths of least action (i.e., expected free energy) from any given state to a preferred or goal state in the future. We will illustrate inductive inference under the RGM above to reproduce expert behaviour.

## Learning expert behaviour

To illustrate inductive (planning as) inference, we equipped the RGM with the prior belief that transitions among the final states (i.e., events) were under its control. This means, *a priori*, it believes that the next event will be on a path to a preferred or rewarded event. These empirical priors are then propagated down the hierarchy to predict the next outcome and drive action, which is selected to minimise variational free energy: see Figure 24 and Equation (1). Because action can only change outcomes, this minimisation reduces to maximising the predictive accuracy of outcomes under allowable actions. In general, this predictive accuracy pertains to proprioceptive or telemetry data reporting the state of the motor plant (Adams et al., 2013; Friston et al., 2011; Lanillos et al., 2021). Here, for simplicity, we used the upper row of visual inputs reporting the location of the paddle. Note the subtlety of this implementation of planning as inference: from the agent's perspective, it does not explicitly represent action or, indeed, anything that can be overtly controlled. At the highest level, it is simply predicting a path to preferred outcomes, which—unsurprisingly from its point of view—unfolds as anticipated (via control as inference). Anthropomorphically, the agent does not know that its own action realised these predictions, or that this realisation was operating at a faster timescale than its plans were being conceived—it just imagines its own future, unaware it is the author of its sensorium.

At this point, one might ask why is inductive inference necessary, given that the RGM can only generate expert play? The answer is that there is a plurality of paths to the next rewarding event. By enabling inductive inference—by specifying events that entail a reward—one can be assured that the paths of least action are selected. One could stop here and illustrate the performance of inductive inference following structure learning. In other words, in principle, one can solve reward-learning problems quickly and efficiently using Bayesian model selection and inductive inference, without recourse to any learning *per se*. However, we can recall the benefits of continual (active) learning in previous sections, which enable the model to generalise to outcomes that have not been encountered before; namely, outcomes that are generated by a degree of unpredictability or randomness.

Figure 27 shows the result of continual learning over 512 exchanges, using the same format as Figure

17). Crucially, we introduced stochasticity into action selection, to confound the otherwise expert play the RGM would exhibit. This was modelled by sampling from the posterior over action, as opposed to selecting the most likely action. In the active inference literature this is enabled by a 'shaky hand' parameter (Parr et al., 2022), p177. In other words, introducing a probabilistic mapping between the agents prescribed movement and the actual execution. This mirrors 'sticky action' in machine learning benchmarks (Machado et al., 2017). Furthermore, we rendered the likelihood mappings slightly uncertain by adding a small concentration parameter of (1/128) to the Dirichlet distributions. In principle, the agent should now be able to learn precise dynamics, under inductive inference. In turn, this should increase the precision of posterior predictions underwriting action selection; enabling more confident play as the agent becomes more experienced. Figure 27 (panel labelled ELBO) suggests that this 'self-confidence' asymptotes after about 400 frames of gameplay.

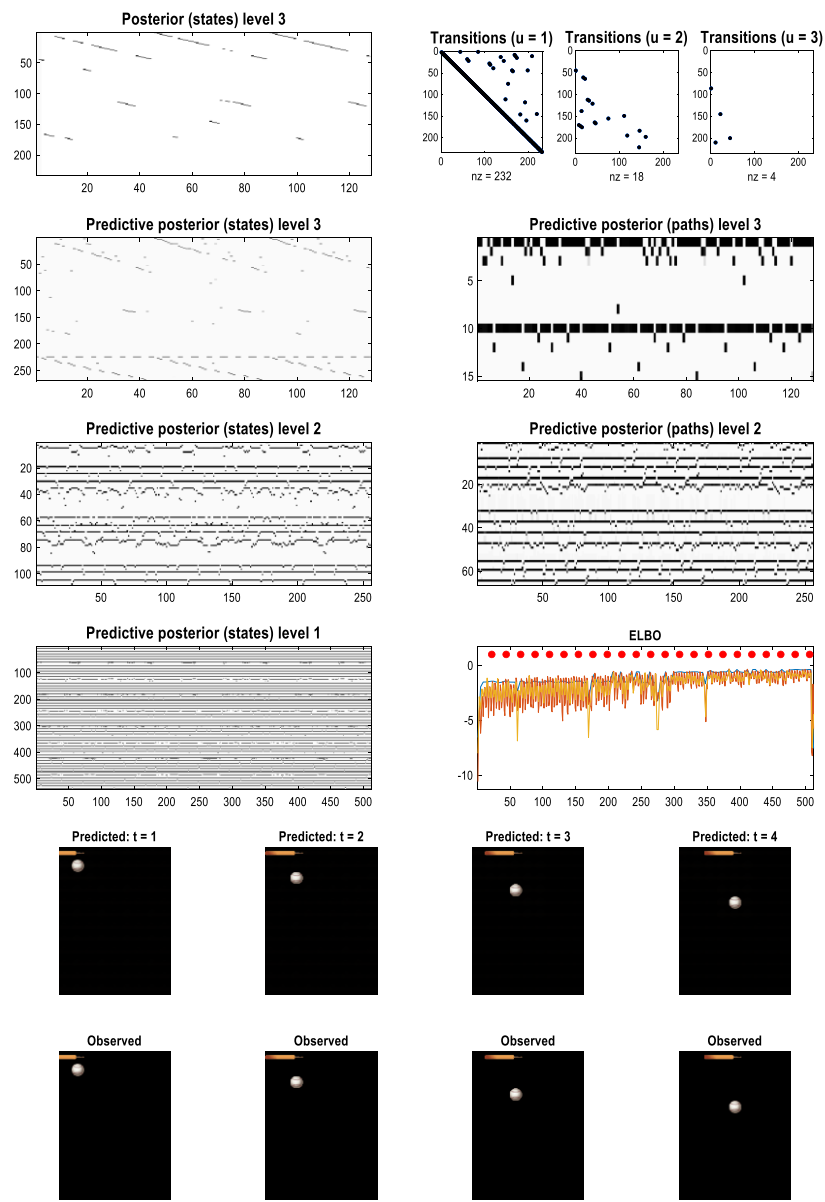

**Figure 27: learning expert play**. This figure summarises the results of fast structure learning after exposure to the selected training set. A sequence of 1024 frames was sampled selectively from 21,280 frames, generated with random paddle movement. The ensuing sequence were learned in about 18 seconds (on a standard PC). This figure follows the format of previous figures, showing three discovered transitions among certain events, corresponding to the alternative ways in which the paddle moved in the training sequence. This sequence has been summarised in terms of transitions among 233 events. The panel labelled ELBO reports the (negative) variational free energy during continual learning of 512 frames of self-generated play. The red dots correspond to (rewarded) hits, while the coloured lines report the ELBO at each level of the RGM. Because the model can only recognise, predict and thereby realise expert play, the implicit agent never misses the ball (in this example). However, it can learn to become more confident in its (realised) predictions as evinced by a gradual increase in the ELBO.

The upper panels of Figure 27 show the posterior predictions emanating from each of the three levels pertaining to states (left column) and paths (right column). Note that there are no posterior predictions over paths at the first level, which is simply generating images in pixel space. The corresponding images for the first four timesteps are shown in the lower panels, in terms of predictions and the stimuli or frames supplied by the game engine. Performance in terms of hits (red dots) and negative variational free energy (ELBO) are shown in the middle panel on the right. Note that the free energy at each successive level fluctuates more slowly because of the renormalisation in time.

## The final game

Figure 28 shows the results of the same simulation using a slightly more complicated game based upon 'Breakout'. In this simplified game, a row of targets was deleted if, and only if, the targets were hit by a returned ball. In this example, the number of frames selected for structure learning—under random action—was doubled to 2,048. Otherwise, the active selection, learning and (inductive) inference were exactly the same as above.

In this example, unlike the pong example, the game is reset. This allowed the introduction of further stochasticity, by replacing the ball to the left or right of the initial location, at random. However, these alternative paths from rewarded states are accommodated within the RGM; provided they are experienced as part of the selected training data. Note that this example introduces a slow change in the nature of events as rows of targets are progressively eliminated. In other words, there are fast dynamics as the ball is returned after hit and a slower succession of episodes as rows of targets are eliminated. This can be seen in the upper left panel of Figure 28 that reports the predictive posteriors from the highest (third) level. Here, certain episodes are repeated regularly but infrequently as the agent revisits the same context, whenever there is a reset.

The additional complexity and stochasticity afforded by this game confound as the agent to the extent it occasionally misses the ball (reported by the gaps in the red dots in the panel labelled ELBO). However, the agent can recover; provided it recognises an event that leads to a reward. As in the previous illustration, Dirichlet counts are slowly accumulated under largely expert play, as reflected in the slow increases in the ELBO as learning progresses.

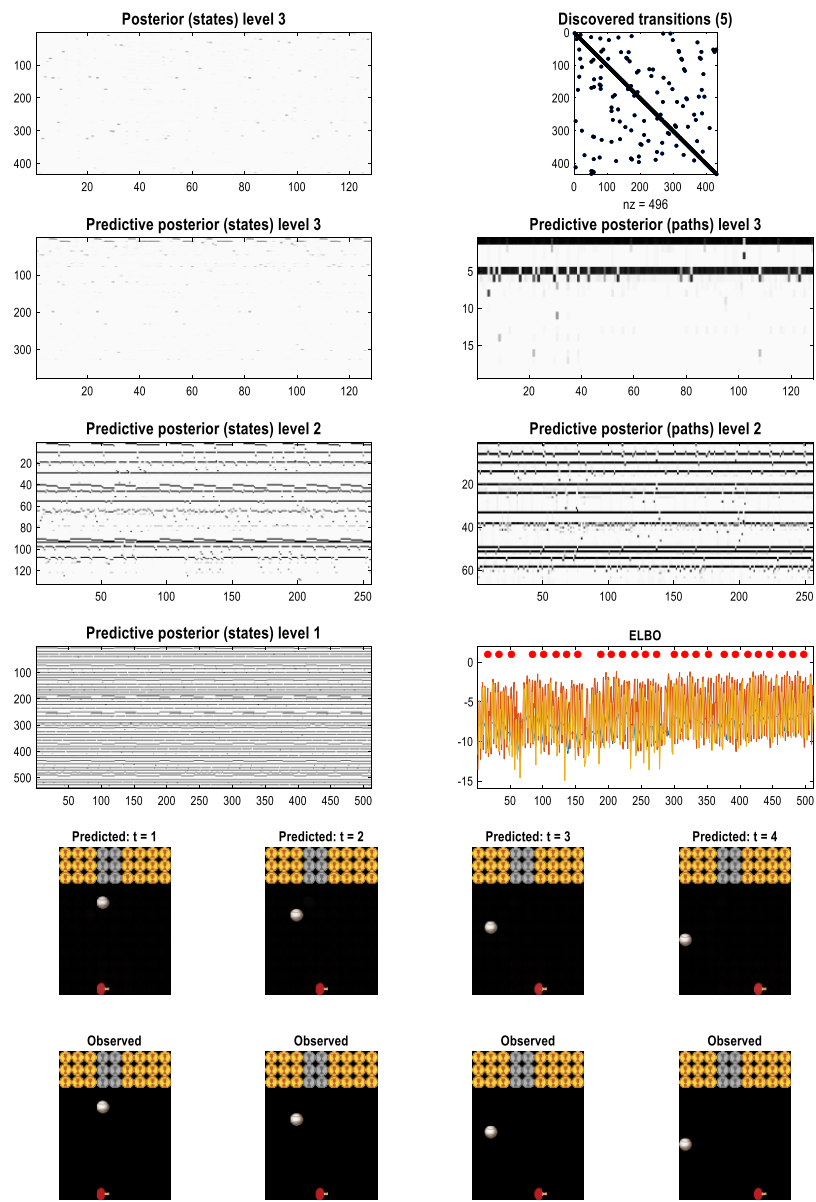

**Figure 28: Breakout**. This figure uses the same format as Figure 27. Here, we repeated the analysis using a slightly more complicated game based upon Breakout (and doubling the number of training frames to 2,048). In this version of Breakout, a reward is recorded whenever the ball hits a row of targets (see lower panels). The row of targets is then removed to expose the underlying row. If a golden target on the final row is hit, the game is reset. Whenever the agent misses a ball—or on reset—the paddle is reset to the centre and the ball appears at a fixed height and randomly selected horizontal location, around the centre. In this example, expert play is confounded by 'sticky action', which means that the movement of the paddle diverges occasionally from the predicted movement. However, the agent recovers quickly and resumes expert play following each miss. This rests upon waiting for a recognisable event that is within the attracting set that leads to a reward. As in the previous example, there is a slow increase in confidence with accumulating Dirichlet counts in the likelihood mappings of the RGM.

Note that because this game has many more configurations than the previous game of Pong, there are more paths among events; here, five such paths (which are shown as discovered transitions by summing over the path dimension of the transition tensor at the final level).

# CONCLUSION

This paper has showcased several applications of a generative model for discrete state-spaces based upon the renormalisation group. The applications—used to explain and illustrate the accompanying belief updating—use exactly the same routines and principles; namely selecting, learning and inverting generative models via minimisation of variational and expected free energy. The appeal to the apparatus of the renormalisation group is relatively straightforward in this setting. This is because active inference is an application of the free energy principle, which itself is a variational principle of least action, whose functional form is conserved over scales or hierarchical levels (Friston, 2019). One can leverage this to provide a belief updating process that is renormalisable over both space and time. The spatial aspect has been foregrounded by application to content or observations in pixel space, motivating the use of spin-block transformations that have proved successful in related applications in physics and machine learning (Hu et al., 2020; Vidal, 2007). Having said this, the restriction to spin-block transformations and pixel spaces could also be regarded as a limitation. In principle, any outcome space could be a candidate for coarse graining, provided some tessellation or partition procedure is at hand—and can be appropriately motivated in terms of conditional independencies: see (Friston et al., 2021b) for a worked example in the context of Markov blankets and the renormalisation group.

RGMs, with quantised paths may, or may not be useful in certain application domains. The applications above illustrate the simplicity and efficiency with which the structure of these models can be learned. Furthermore, their discrete nature lends the requisite belief-updating a straightforward form; resting largely on sum-product operators, which can be exact in the sense of Bayesian inference. In turn, this enables efficient forms of planning; for example, inductive inference (Friston et al., 2023c). The downside of these models is that their parameterisation may not be apt for complex system modelling—and associated scenario or intervention modelling—that is usually predicated on continuous state-space models. For example, dynamic causal models are generally parameterised in terms of rate constants in the context of ordinary or stochastic differential equations. This means that there is a biophysical interpretation to the model parameters that can be lost following discretisation or, indeed, discrete switching among dynamical systems. Having said this, it is always possible to convert a continuous state-space model into a master equation, specifying transitions among discrete states: e.g., compartmental models of the sort used in biochemistry and epidemiology (Seifert, 2005; Wang, 2009; Wolkenhauer et al., 2004). This speaks to the possibility of learning discrete state-space models using renormalisation procedures—and then replacing the Dirichlet parameterisation with forms parameterised in terms of rate constants and accompanying probability fluxes among compartments or states.

Further limitations of the approach described above rest on the efficiency of the accompanying methods. This efficiency is both a gift and a burden. Clearly, in terms of sample efficiency and compression, the schemes described above are designed to outperform conventional learning schemes. This follows because the model selection procedure is constructive—as opposed to the reductive approach of Bayesian model reduction and related procedures—enabling the self-assembly of a minimally expressive model to recognise, generate and predict the content at hand. This should be contrasted with the reductive approaches that start with a large, overly expressive model that is then reduced or pruned, minimising complexity to recover predictive validity and generalisation.

However, in committing to a minimally complex—maximally efficient—model structure, one is necessarily committed to the kind of data used for model selection and learning. This means such models are necessarily brittle, in the sense that they will not recognise or respond to events they have not previously encountered. A useful heuristic here is learning to ride a bike: in this kind of procedural learning, we learn from a starting position and accumulate successive cycles of peddling, until we avoid falling completely. In other words, we retain only those experiences characteristic of successful bike riding, until we can ride fluently. However, we would not be able to ride a bike starting from any arbitrary state, e.g., having fallen over. We would have to return to a known starting position to recover our flow. Flow is meant in manifold senses here: the fluency associated with being in the flow (Parvizi-Wayne et al., 2024), control flow (Fields et al., 2023a, b), dynamical flow (Huys et al., 2014; Rabinovich et al., 2012) and, of course, RG-flow (Hu et al., 2020).

## Acknowledgements

KF is supported by funding for the Wellcome Centre for Human Neuroimaging (Ref: 205103/Z/16/Z), a Canada-UK Artificial Intelligence Initiative (Ref: ES/T01279X/1) and the European Union's Horizon 2020 Framework Programme for Research and Innovation under the Specific Grant Agreement No. 945539 (Human Brain Project SGA3). TP is supported by an NIHR Academic Clinical Fellowship (ref: ACF-2023-13-013).

## Disclosure statement

The authors have no disclosures or conflict of interest.

## References

Adams, R.A., Shipp, S., Friston, K.J., 2013. Predictions not commands: active inference in the motor system. Brain Struct Funct. 218, 611-643.

Alpers, G.W., Gerdes, A.B.M., 2007. Here is looking at you: emotional faces predominate in binocular rivalry. Emotion 7, 495-506.

Angelucci, A., Bullier, J., 2003. Reaching beyond the classical receptive field of V1 neurons: horizontal or feedback axons? Journal of physiology, Paris 97, 141-154.

Attias, H., 2003. Planning by Probabilistic Inference, Proc. of the 9th Int. Workshop on Artificial Intelligence and Statistics.

Ay, N., Bertschinger, N., Der, R., Guttler, F., Olbrich, E., 2008. Predictive information and explorative behavior of autonomous robots. European Physical Journal B 63, 329-339.

Bach, S., Binder, A., Montavon, G., Klauschen, F., Muller, K.R., Samek, W., 2015. On Pixel-Wise Explanations for Non-Linear Classifier Decisions by Layer-Wise Relevance Propagation. Plos One 10.

Bagaev, D., de Vries, B., 2021. Reactive Message Passing for Scalable Bayesian Inference, p. arXiv:2112.13251.

Barlow, H., 1961. Possible principles underlying the transformations of sensory messages, in: Rosenblith, W. (Ed.), Sensory Communication. MIT Press, Cambridge, MA, pp. 217-234.

Bastos, A.M., Vezoli, J., Bosman, C.A., Schoffelen, J.M., Oostenveld, R., Dowdall, J.R., De Weerd, P., Kennedy, H., Fries, P., 2015. Visual areas exert feedforward and feedback influences through distinct frequency channels. Neuron 85, 390-401.

Beal, M.J., 2003. Variational Algorithms for Approximate Bayesian Inference. PhD. Thesis, University College London.

Berger, J.O., 2011. Statistical decision theory and Bayesian analysis. Springer, New York; London.

Botvinick, M., Toussaint, M., 2012. Planning as inference. Trends in Cognitive Sciences 16, 485-488.

Braun, D.A., Ortega, P.A., Theodorou, E., Schaal, S., 2011. Path integral control and bounded rationality, Adaptive Dynamic Programming And Reinforcement Learning (ADPRL), 2011 IEEE Symposium on. IEEE, Paris, pp. 202 - 209.

Cardy, J.L., 2015. Scaling and renormalization in statistical physics.

Cugliandolo, L.F., Lecomte, V., 2017. Rules of calculus in the path integral representation of white noise Langevin equations: the Onsager-Machlup approach. Journal of Physics a-Mathematical and Theoretical 50, 345001.

Da Costa, L., Parr, T., Sajid, N., Veselic, S., Neacsu, V., Friston, K., 2020. Active inference on discrete state-spaces: A synthesis. J Math Psychol 99, 102447.

Dauwels, J., 2007. On Variational Message Passing on Factor Graphs, 2007 IEEE International Symposium on Information Theory, pp. 2546-2550.

Deyle, E.R., Sugihara, G., 2011. Generalized theorems for nonlinear state space reconstruction. PLoS One 6, e18295.

Feldman, A.G., 2009. New insights into action-perception coupling. Experimental Brain Research 194, 39-58.

Fields, C., Fabrocini, F., Friston, K., Glazebrook, J.F., Hazan, H., Levin, M., Marcianò, A., 2023a. Control Flow in Active Inference Systems—Part I: Classical and Quantum Formulations of Active Inference. IEEE Transactions on Molecular, Biological and Multi-Scale Communications 9, 235-245.

Fields, C., Fabrocini, F., Friston, K., Glazebrook, J.F., Hazan, H., Levin, M., Marcianò, A., 2023b. Control Flow in Active Inference Systems—Part II: Tensor Networks as General Models of Control Flow. IEEE Transactions on Molecular, Biological and Multi-Scale Communications 9, 246-256.

Friston, K., 2011. What is optimal about motor control? Neuron 72, 488-498.

Friston, K., 2019. A free energy principle for a particular physics, eprint arXiv:1906.10184.

Friston, K., Da Costa, L., Sakthivadivel, D.A.R., Heins, C., Pavliotis, G.A., Ramstead, M., Parr, T., 2023a. Path integrals, particular kinds, and strange things. Physics of Life Reviews 47, 35-62.

Friston, K., Heins, C., Ueltzhoffer, K., Da Costa, L., Parr, T., 2021a. Stochastic Chaos and Markov Blankets. Entropy (Basel) 23, 1220.

Friston, K., Kiebel, S., 2009. Predictive coding under the free-energy principle. Philosophical transactions of the Royal Society of London. Series B, Biological sciences 364, 1211-1221.

Friston, K., Mattout, J., Kilner, J., 2011. Action understanding and active inference. Biological cybernetics 104, 137-160.

Friston, K., Parr, T., Zeidman, P., 2018. Bayesian model reduction. arXiv preprint arXiv:1805.07092.

Friston, K., Penny, W., 2011. Post hoc Bayesian model selection. Neuroimage 56, 2089-2099.

Friston, K., Rigoli, F., Ognibene, D., Mathys, C., Fitzgerald, T., Pezzulo, G., 2015. Active inference and epistemic value. Cogn Neurosci, 1-28.

Friston, K.J., Da Costa, L., Tschantz, A., Kiefer, A., Salvatori, T., Neacsu, V., Koudahl, M., Heins, C., Sajid, N., Markovic, D., Parr, T., Verbelen, T., Buckley, C.L., 2023b. Supervised structure learning, p. arXiv:2311.10300.

Friston, K.J., Fagerholm, E.D., Zarghami, T.S., Parr, T., Hipolito, I., Magrou, L., Razi, A., 2021b. Parcels and particles: Markov blankets in the brain. Network neuroscience (Cambridge, Mass.) 5, 211-251.

Friston, K.J., Fagerholm, E.D., Zarghami, T.S., Parr, T., Hipólito, I., Magrou, L., Razi, A., 2021c. Parcels and particles: Markov blankets in the brain. Network Neuroscience 5, 211-251.

Friston, K.J., Frith, C.D., 2015. Active inference, communication and hermeneutics. Cortex; a journal devoted to the study of the nervous system and behavior 68, 129-143.

Friston, K.J., Parr, T., de Vries, B., 2017a. The graphical brain: Belief propagation and active inference. Network neuroscience (Cambridge, Mass.) 1, 381-414.

Friston, K.J., Parr, T., Zeidman, P., Razi, A., Flandin, G., Daunizeau, J., Hulme, O.J., Billig, A.J., Litvak, V., Price, C.J., Moran, R.J., Lambert, C., 2020. Second waves, social distancing, and the spread of COVID-19 across the USA. Wellcome Open Res 5, 103.

Friston, K.J., Rosch, R., Parr, T., Price, C., Bowman, H., 2017b. Deep temporal models and active inference. Neuroscience and biobehavioral reviews 77, 388-402.

Friston, K.J., Salvatori, T., Isomura, T., Tschantz, A., Kiefer, A., Verbelen, T., Koudahl, M., Paul, A., Parr, T., Razi, A., Kagan, B., Buckley, C.L., Ramstead, M.J.D., 2023c. Active Inference and Intentional Behaviour, p. arXiv:2312.07547.

Friston, K.J., Stephan, K., Li, B.J., Daunizeau, J., 2010. Generalised Filtering. Mathematical Problems in Engineering 2010, 621670.

Ghavamzadeh, M., Mannor, S., Pineau, J., Tamar, A., 2016. Bayesian Reinforcement Learning: A Survey. arXiv arXiv:1609.04436.

Goodale, M.A., Westwood, D.A., Milner, A.D., 2004. Two distinct modes of control for object-directed action. Prog Brain Res 144, 131-144.

Gros, C., 2009. Cognitive Computation with Autonomously Active Neural Networks: An Emerging Field. Cognitive Computation 1, 77-90.

Hasson, U., Yang, E., Vallines, I., Heeger, D.J., Rubin, N., 2008. A hierarchy of temporal receptive windows in human cortex. J Neurosci 28, 2539-2550.

Hewitt, C., Bishop, P., Steiger, R., 1973. A universal modular ACTOR formalism for artificial intelligence, Proceedings of the 3rd international joint conference on Artificial intelligence. Morgan Kaufmann Publishers Inc., Stanford, USA, pp. 235–245.

Higgins, I., Chang, L., Langston, V., Hassabis, D., Summerfield, C., Tsao, D., Botvinick, M., 2021. Unsupervised deep learning identifies semantic disentanglement in single inferotemporal face patch neurons. Nat Commun 12, 6456.

Hochstein, S., Ahissar, M., 2002. View from the top: hierarchies and reverse hierarchies in the visual system. Neuron 36, 791-804.

Howard, R., 1966. Information Value Theory. IEEE Transactions on Systems, Science and Cybernetics SSC-2, 22-26.

Hu, H.-y., Wu, D., You, Y.-Z., Olshausen, B.A., Chen, Y., 2020. RG-Flow: a hierarchical and explainable flow model based on renormalization group and sparse prior. Machine Learning: Science and Technology 3.

Huys, R., Perdikis, D., Jirsa, V.K., 2014. Functional architectures and structured flows on manifolds: a dynamical framework for motor behavior. Psychol Rev 121, 302-336.

Itti, L., Baldi, P., 2009. Bayesian Surprise Attracts Human Attention. Vision Res. 49, 1295-1306.

Kaiser, L., Babaeizadeh, M., Milos, P., Osinski, B., Campbell, R.H., Czechowski, K., Erhan, D., Finn, C., Kozakowski, P., Levine, S., Mohiuddin, A., Sepassi, R., Tucker, G., Michalewski, H., 2019. Model-Based Reinforcement Learning for Atari, p. arXiv:1903.00374.

Kaluza, P., Meyer-Ortmanns, H., 2010. On the role of frustration in excitable systems. Chaos 20, 043111.

Kappen, H.J., Gomez, V., Opper, M., 2012. Optimal control as a graphical model inference problem. Machine Learning 87, 159-182.

Kerr, W.C., Graham, A.J., 2000. Generalized phase space version of Langevin equations and associated Fokker-Planck equations. European Physical Journal B 15, 305-311.

Kersten, D., Mamassian, P., Yuille, A., 2004. Object perception as Bayesian inference. Annual review of psychology 55, 271-304.

Kiebel, S.J., Daunizeau, J., Friston, K.J., 2008. A hierarchy of time-scales and the brain. PLoS Comput Biol 4, e1000209.

Klimovskaia, A., Lopez-Paz, D., Bottou, L., Nickel, M., 2020. Poincare maps for analyzing complex hierarchies in single-cell data. Nat Commun 11, 2966.

Klyubin, A.S., Polani, D., Nehaniv, C.L., 2005. Empowerment: A Universal Agent-Centric Measure of Control. In Proc. CEC 2005. IEEE 1, 128-135.

Lanillos, P., Meo, C., Pezzato, C., Meera, A.A., Baioumy, M., Ohata, W., Tschantz, A., Millidge, B., Wisse, M., Buckley, C.L., Tani, J., 2021. Active Inference in Robotics and Artificial Agents: Survey and Challenges, p. arXiv:2112.01871.

LeCun, Y., Cortes, C., 2005. The mnist database of handwritten digits.

Lillicrap, T.P., Hunt, J.J., Pritzel, A., Heess, N., Erez, T., Tassa, Y., Silver, D., Wierstra, D., 2015. Continuous control with deep reinforcement learning. arXiv e-prints, arXiv:1509.02971.

Lin, H.W., Tegmark, M., Rolnick, D., 2017. Why Does Deep and Cheap Learning Work So Well? Journal of Statistical Physics 168, 1223-1247.

Linderman, S.W., Miller, A.C., Adams, R.P., Blei, D.M., Paninski, L., Johnson, M.J., 2016. Recurrent switching linear dynamical systems. eprint arXiv:1610.08466, 1-15.

Lindley, D.V., 1956. On a Measure of the Information Provided by an Experiment. Annals of Mathematical Statistics 27, 986-1005.

Linsker, R., 1990. Perceptual Neural Organization - Some Approaches Based on Network Models and Information-Theory. Annual Review of Neuroscience 13, 257-281.

Livingstone, M., Hubel, D., 1988. Segregation of form, color, movement, and depth: anatomy, physiology, and perception. Science 240, 740-749.

Lorenz, E.N., 1963. Deterministic Nonperiodic Flow. Journal of the Atmospheric Sciences 20, 130-141.

Ma, Y., Tan, Q.J., Yuan, R.S., Yuan, B., Ao, P., 2014. Potential Function in a Continuous Dissipative Chaotic System: Decomposition Scheme and Role of Strange Attractor. International Journal of Bifurcation and Chaos 24, 1450015.

Machado, M.C., Bellemare, M.G., Talvitie, E., Veness, J., Hausknecht, M., Bowling, M., 2017. Revisiting the Arcade Learning Environment: Evaluation Protocols and Open Problems for General Agents, p. arXiv:1709.06009.

MacKay, D.J.C., 2003. Information Theory, Inference and Learning Algorithms. Cambridge University Press, Cambridge.

MacKay, D.J.C., Peto, L.C.B., 2008. A hierarchical Dirichlet language model. Natural Language Engineering 1, 289-308.

Mansell, W., 2011. Control of perception should be operationalized as a fundamental property of the nervous system. Top Cogn Sci 3, 257-261.

Marković, D., Reiter, A.M.F., Kiebel, S.J., 2022. Revealing human sensitivity to a latent temporal structure of changes. Front Behav Neurosci 16, 962494.

Mehta, P., Schwab, D.J., 2014. An exact mapping between the Variational Renormalization Group and Deep Learning, p. arXiv:1410.3831.

Muckli, L., De Martino, F., Vizioli, L., Petro, L.S., Smith, F.W., Ugurbil, K., Goebel, R., Yacoub, E., 2015. Contextual Feedback to Superficial Layers of V1. Current Biology 25, 2690-2695.

Olier, I., Trujillo-Barreto, N.J., El-Deredy, W., 2013. A switching multi-scale dynamical network model of EEG/MEG. NeuroImage 83, 262-287.

Olshausen, B.A., Field, D.J., 1996. Emergence of simple-cell receptive field properties by learning a sparse code for natural images. Nature 381, 607-609.

Ørstavik, S., Stark, J., 1998. Reconstruction and cross-prediction in coupled map lattices using spatio-temporal embedding techniques. Physics Letters A 247, 145-160.

Ortega, P.A., Braun, D.A., 2013. Thermodynamics as a theory of decision-making with information-processing costs. Proc. R. Soc. A 469 2153.

Parr, T., Pezzulo, G., Friston, K.J., 2022. Active Inference: The Free Energy Principle in Mind, Brain, and Behavior. MIT Press, Cambridge.

Parvizi-Wayne, D., Sandved-Smith, L., Pitliya, R.J., Limanowski, J., Tufft, M.R.A., Friston, K.J., 2024. Forgetting ourselves in flow: an active inference account of flow states and how we experience ourselves within them. Frontiers in Psychology 15.

Pefkou, M., Arnal, L.H., Fontolan, L., Giraud, A.L., 2017. theta-Band and beta-Band Neural Activity Reflects Independent Syllable Tracking and Comprehension of Time-Compressed Speech. J Neurosci 37, 7930-7938.

Poland, D., 1993. Cooperative Catalysis and Chemical Chaos - a Chemical-Model for the Lorenz Equations. Physica D 65, 86-99.

Rabinovich, M.I., Afraimovich, V.S., Bick, C., Varona, P., 2012. Information flow dynamics in the brain. Phys Life Rev 9, 51-73.

Rabinovich, M.I., Afraimovich, V.S., Varona, P., 2010. Heteroclinic binding. Dynamical Systems-an International Journal 25, 433-442.

Ramstead, M.J.D., Sakthivadivel, D.A.R., Heins, C., Koudahl, M., Millidge, B., Da Costa, L., Klein, B., Friston, K.J., 2022. On Bayesian Mechanics: A Physics of and by Beliefs, p. arXiv:2205.11543.

Sakthivadivel, D.A.R., 2022. Weak Markov Blankets in High-Dimensional, Sparsely-Coupled Random Dynamical Systems, p. arXiv:2207.07620.

Sanchez, E.H., Serrurier, M., Ortner, M., 2019. Learning Disentangled Representations via Mutual Information Estimation, p. arXiv:1912.03915.

Savage, L.J., 1954. The Foundations of Statistics. Wiley, New York.

Schmidhuber, J., 1991. Curious model-building control systems. In Proc. International Joint Conference on Neural Networks, Singapore. IEEE 2, 1458–1463.

Schwabl, F., 2002. Phase Transitions, Scale Invariance, Renormalization Group Theory, and Percolation, Statistical Mechanics. Springer Berlin Heidelberg, Berlin, Heidelberg, pp. 327-404.

Schwenzer, M., Ay, M., Bergs, T., Abel, D., 2021. Review on model predictive control: an engineering perspective. International Journal of Advanced Manufacturing Technology 117, 1327-1349.

Seifert, U., 2005. Entropy production along a stochastic trajectory and an integral fluctuation theorem. Phys Rev Lett 95, 040602.

Sengupta, B., Friston, K., 2018. How Robust are Deep Neural Networks? arXiv arXiv:1804.11313.

Simoncelli, E.P., Olshausen, B.A., 2001. Natural image statistics and neural representation. Annu Rev Neurosci 24, 1193-1216.

Smith, R., Schwartenbeck, P., Parr, T., Friston, K.J., 2020. An Active Inference Approach to Modeling Structure Learning: Concept Learning as an Example Case. Frontiers in computational neuroscience 14, 41.

Still, S., Precup, D., 2012. An information-theoretic approach to curiosity-driven reinforcement learning. Theory in biosciences = Theorie in den Biowissenschaften 131, 139-148.

Takens, F., 1980. Detecting strange attractors in turbulence. Rijksuniversiteit Groningen. Mathematisch Instituut, Groningen.

Tani, J., Nolfi, S., 1999. Learning to perceive the world as articulated: an approach for hierarchical learning in sensory-motor systems. Neural Networks 12, 1131-1141.

Teh, Y.W., Jordan, M.I., Beal, M.J., Blei, D.M., 2006. Hierarchical Dirichlet processes. Journal of the American Statistical Association 101, 1566-1581.

Tenenbaum, J.B., Kemp, C., Griffiths, T.L., Goodman, N.D., 2011. How to grow a mind: statistics, structure, and abstraction. Science 331, 1279-1285.

Tervo, D.G.R., Tenenbaum, J.B., Gershman, S.J., 2016. Toward the neural implementation of structure learning. Curr Opin Neurobiol 37, 99-105.

Theodorou, E.A., Buchli, J., Schaal, S., 2010. A Generalized Path Integral Control Approach to Reinforcement Learning. Journal of Machine Learning Research 11, 3137-3181.

Tipping, M.E., 2001. Sparse Bayesian learning and the relevance vector machine. Journal of Machine Learning Research 1, 211-244.

Todorov, E., Jordan, M.I., 2002. Optimal feedback control as a theory of motor coordination. Nature Neuroscience 5, 1226-1235.

Tomasello, M., 2016. Cultural Learning Redux. Child development 87, 643-653.

Ungerleider, L.G., Mishkin, M., 1982. Two cortical visual systems, in: .Ingle, D., Goodale, M.A., Mansfield, R.J.W. (Eds.), Analysis of Visual Behavior. MIT Press, Cambridge, MA, pp. 549-586.

van den Broek, J.L., Wiegerinck, W.A.J.J., Kappen, H.J., 2010. Risk-sensitive path integral control. UAI 6, 1–8.

Van Dijk, S.G., Polani, D., 2013. Informational Constraints-Driven Organization in Goal-Directed Behavior. Advances in Complex Systems 16, 1350016.

Vidal, G., 2007. Entanglement renormalization. Physical Review Letters 99.

Wang, X.M., 2009. From Dirac Notation to Probability Bracket Notation: Time Evolution and Path Integral under Wick Rotations. ArXiv e-prints.

Watkins, C.J.C.H., Dayan, P., 1992. Q-Learning. Machine Learning 8, 279-292.

Watson, J.D., Onorati, E., Cubitt, T.S., 2022. Uncomputably complex renormalisation group flows. Nature Communications 13, 7618.

Winn, J., Bishop, C.M., 2005. Variational message passing. Journal of Machine Learning Research 6, 661-694.

Wolkenhauer, O., Ullah, M., Kolch, W., Cho, K.H., 2004. Modeling and simulation of intracellular dynamics: choosing an appropriate framework. IEEE Trans Nanobioscience 3, 200-207.

Yildiz, I.B., von Kriegstein, K., Kiebel, S.J., 2013. From birdsong to human speech recognition: bayesian inference on a hierarchy of nonlinear dynamical systems. PLoS Comput Biol 9, e1003219.

Zeki, S., Shipp, S., 1988. The functional logic of cortical connections. Nature 335, 311-317.